%% file: icml20.tex
\documentclass{article}

\usepackage{url}
\input{macros}

\usepackage[accepted]{icml2020}

\begin{document}

\twocolumn[
\icmltitlerunning{Evaluating Lossy Compression Rates of Deep Generative Models}
\icmltitle{Evaluating Lossy Compression Rates of Deep Generative Models}

\icmlsetsymbol{equal}{*}

\begin{icmlauthorlist}
\icmlauthor{Sicong Huang}{equal,uoft,vector,borealis}
\icmlauthor{Alireza Makhzani}{equal,uoft,vector}
\icmlauthor{Yanshuai Cao}{borealis}
\icmlauthor{Roger Grosse}{uoft,vector}
\end{icmlauthorlist}

\icmlaffiliation{uoft}{University of Toronto}
\icmlaffiliation{vector}{Vector Institute for Artificial Intelligence}
\icmlaffiliation{borealis}{Borealis AI}

\icmlcorrespondingauthor{Alireza Makhzani, Roger Grosse}{makhzani, rgrosse@cs.toronto.edu}

\icmlkeywords{Deep Learning, Generative Models, Information Theory, Rate Distortion Theory}

\vskip 0.3in
]

\printAffiliationsAndNotice{\icmlEqualContribution}

\begin{abstract}
The field of deep generative modeling has succeeded in producing astonishingly realistic-seeming images and audio, but quantitative evaluation remains a challenge. Log-likelihood is an appealing metric due to its grounding in statistics and information theory, but it can be challenging to estimate for implicit generative models, and scalar-valued metrics give an incomplete picture of a model's quality. In this work, we propose to use rate distortion (RD) curves to evaluate and compare deep generative models. While estimating RD curves is seemingly even more computationally demanding than log-likelihood estimation, we show that we can approximate the entire RD curve using nearly the same computations as were previously used to achieve a single log-likelihood estimate. We evaluate lossy compression rates of VAEs, GANs, and adversarial autoencoders (AAEs) on the MNIST and CIFAR10 datasets. Measuring the entire RD curve gives a more complete picture than scalar-valued metrics, and we arrive at a number of insights not obtainable from log-likelihoods alone.
\end{abstract}

\section{Introduction}

Generative models of images represent one of the most exciting areas of rapid progress of AI~\citep{brock2018large,karras2018style,karras2018progressive}. However, evaluating the performance of generative models remains a significant challenge. Many of the most successful models, most notably Generative Adversarial Networks (GANs)~\citep{goodfellow2014generative}, are \emph{implicit generative models} for which computation of log-likelihoods is intractable or even undefined. Evaluation typically focuses on metrics such as the Inception score~\citep{salimans2016improved} or the Fr\'{e}chet Inception Distance (FID)~\citep{fid}, which do not have nearly the same degree of theoretical underpinning as likelihood-based metrics. 

Log-likelihoods are one of the most important measures of generative models. Their utility is evidenced by the fact that likelihoods (or equivalent metrics such as perplexity or bits-per-dimension) are reported in nearly all cases where it's convenient to compute them. Unfortunately, computation of log-likelihoods for implicit generative models remains a difficult problem. Furthermore, log-likelihoods have important conceptual limitations. For continuous inputs in the image domain, the metric is often dominated by the fine-grained distribution over pixels rather than the high-level structure. For models with low-dimensional support, one needs to assign an observation model, such as (rather arbitrary) isotropic Gaussian noise~\citep{wu2016quantitative}.
Lossless compression metrics for GANs often give absurdly large bits-per-dimension (e.g.~$10^{14}$) which fails to reflect the true performance of the model~\citep{grover2018flow,danihelka2017comparison}.
See~\citet{theis2015note} for more discussion of limitations of likelihood-based evaluation.

Typically, one is not interested in describing the pixels of an image directly, and it would be sufficient to generate images close to the true data distribution in some metric such as Euclidean distance. For this reason, there has been much interest in Wasserstein distance as a criterion for generative models, since the measure exploits precisely this metric structure~\citep{arjovsky2017wasserstein,gulrajani2017improved,salimans2018improving}. 
However, Wasserstein distance remains difficult to approximate, and hence it is not routinely used to evaluate generative models. 

We aim to achieve the best of both worlds by measuring \emph{lossy compression} rates of deep generative models. In particular, we aim to estimate the rate distortion function, which measures the number of bits required to match a distribution to within a given distortion. Like Wasserstein distance, it can exploit the metric structure of the observation space, but like log-likelihoods, it connects to the rich literature of probabilistic and information theoretic analysis of generative models. By focusing on different parts of the rate distortion curve, one can achieve different tradeoffs between the description length and the fidelity of reconstruction --- thereby fixing the problem whereby lossless compression focuses on the details at the expense of high-level structure. The lossy compression perspective has the further advantage that the distortion metric need not have a probabilistic interpretation; hence, one is free to use more perceptually valid distortion metrics such as structural similarity (SSIM)~\citep{wang2004image} or distances in a learned feature space~\citep{huang2018an}.

Algorithmically, computing rate distortion functions raises similar challenges to estimating log-likelihoods. We show that the rate distortion curve can be computed by finding the normalizing constants of a family of unnormalized probability distributions over the noise variables $\mathbf{z}$. Interestingly, when the distortion metric is squared error, these distributions correspond to the posterior distributions of $\mathbf{z}$ for Gaussian observation models with different variances; hence, the rate distortion analysis generalizes the evaluation of log-likelihoods with Gaussian observation models. 

Annealed Importance Sampling (AIS)~\citep{neal2001annealed} is currently the most effective general-purpose method for estimating normalizing constants in high dimensions, and was used by~\citet{wu2016quantitative} to compare log-likelihoods of a variety of implicit generative models. The algorithm is based on gradually interpolating between a tractable initial distribution and an intractable target distribution. We show that when AIS is used to estimate log-likelihoods under a Gaussian observation model, the sequence of intermediate distributions corresponds to precisely the distributions needed to compute the rate distortion curve. We prove that the AIS estimate of the rate distortion curve is an upper bound on the \emph{entire} rate distortion curve. Furthermore, the tightness of the bound can be validated on simulated data using bidirectional Monte Carlo (BDMC)~\citep{grosse2015sandwiching, wu2016quantitative}. Hence, we can approximate the entire rate distortion curve for roughly the same computational cost as a \emph{single log-likelihood estimate}. 

We use our rate distortion approximations to study a variety of variational autoencoders (VAEs)~\citep{kingma2013auto}, GANs and adversarial autoencoders (AAE)~\citep{makhzani2015adversarial}, and arrive at a number of insights not obtainable from log-likelihoods alone. For instance, we observe that VAEs and GANs have different rate distortion tradeoffs: While VAEs with larger code size can generally achieve better lossless compression rates, their performances drop at lossy compression in the low-rate regime. Conversely, expanding the capacity of GANs appears to bring substantial reductions in distortion at the high-rate regime without any corresponding deterioration in quality in the low-rate regime. We find that increasing the capacity of GANs by increasing the code size (width) has a qualitatively different effect on the rate distortion tradeoffs than increasing the depth. We also find that different GAN variants with the same code size achieve nearly identical RD curves, and that the code size dominates the performance differences between GANs.

\section{Background}
\label{sec:background}

\subsection{Annealed Importance Sampling} \label{sec:ais}

Annealed importance sampling (AIS)~\citep{neal2001annealed} is a Monte Carlo algorithm based on constructing a sequence of $n+1$ distributions $p_k(\mathbf{z})={\tilde{p}_k(\mathbf{z})\over Z_k}$, where $k\in\{0,\ldots,n\}$, between a tractable initial distribution $p_0(\mathbf{z})$ and the intractable target distribution $p_n(\mathbf{z})$.
At the $k$-th state ($0\leq k \leq n$), the forward distribution $q_f$ and the un-normalized backward distribution $\tilde{q}_b$ are

\vspace{-.5cm}
\begin{footnotesize}
\begin{align}
q_f(\mathbf{z}_0, \ldots , \mathbf{z}_k)&=p_0(\mathbf{z}_0)\mathcal{T}_0(\mathbf{z}_1|\mathbf{z}_0)\ldots \mathcal{T}_{k-1}(\mathbf{z}_k|\mathbf{z}_{k-1}),\\
\tilde{q}_b(\mathbf{z}_0, \ldots , \mathbf{z}_k)&=\tilde{p}_k(\mathbf{z}_k)\tilde{\mathcal{T}}_{k-1}(\mathbf{z}_{k-1}|\mathbf{z}_k)\ldots\tilde{\mathcal{T}}_{0}(\mathbf{z}_0|\mathbf{z}_{1}),
\end{align}
\end{footnotesize}
\vspace{-.5cm}

where $\mathcal{T}_k$ is an MCMC kernel that leaves $p_k(\mathbf{z})$ invariant; and $\tilde{\mathcal{T}}_k$ is its reverse kernel.
We run $M$ independent AIS chains, numbered $i=1, \ldots, M$. Let $\mathbf{z}^i_k$ be the $k$-th state of the $i$-th chain.
The importance weights $w^{i}_k$ and normalized importance weights $\tilde{w}^{i}_k$ are
\begin{footnotesize}
\begin{align}\label{eq:ais_weights}
w^{i}_k &= 
\frac{\tilde{q}_b(\mathbf{z}^i_1, \ldots, \mathbf{z}^i_k)}{q_f(\mathbf{z}^i_1, \ldots, \mathbf{z}^i_k)}=
{{\tilde{p}_1(\mathbf{z}^i_1)}\over{p_0(\mathbf{z}^i_1)}}
{{\tilde{p}_2(\mathbf{z}^i_2)}\over{\tilde{p}_1(\mathbf{z}^i_2)}}\ldots
{{\tilde{p}_k(\mathbf{z}^i_k)}\over{\tilde{p}_{k-1}(\mathbf{z}^i_{k})}} \\
\tilde{w}^{i}_k &= \frac{{w}^{i}_k}{\sum_{i=1}^{M} {w}^{i}_k}.
\end{align}
\end{footnotesize}
At the $k$-th step, an unbiased estimate of the partition function of $p_k(\mathbf{z})$ can be found using ${\hat{Z}_k} = {1 \over M} \sum_{i=1}^{M} w^{i}_k$.

At the $k$-th step, we define the \emph{AIS distribution} $q_{k}^{\text{AIS}}(\mathbf{z})$ as the distribution obtained by first sampling $\mathbf{z}^1_k, \ldots, \mathbf{z}^M_k$ from the $M$ parallel chains using the forward distribution $q_f(\mathbf{z}^i_1, \ldots, \mathbf{z}^i_M)$, and then re-sampling these samples based on $\tilde{w}^{i}_k$. More formally, the AIS distribution is defined as follows:

\vspace{-.7cm}
\begin{footnotesize}
\begin{align}
q_{k}^{\text{AIS}}(\mathbf{z}) = \mathbb{E}_{\prod_{i=1}^M q_f(\mathbf{z}^i_1, \ldots, \mathbf{z}^i_k)}[\sum_{i=1}^M \tilde{w}^{i}_k\delta(\mathbf{z}-\mathbf{z}^i_k)].
\end{align}
\end{footnotesize}
\vspace{-.7cm}

\subsection{Bidirectional Monte Carlo.} 

We know that the AIS log partition function estimate $\log \hat{Z}$ is a \emph{stochastic lower bound} on $\log Z$ (Jensen's inequality). As the result, using the forward AIS distribution as the proposal distribution results in a lower bound on the data log-likelihood. By running AIS in reverse, however, we obtain an upper bound on $\log Z$. However, in order to run the AIS in reverse, we need exact samples from the true posterior, which is only possible on the simulated data. The combination of the AIS lower bound and upper bound on the log partition function is called \emph{bidirectional Monte Carlo} (BDMC), and the gap between these bounds is called the \emph{BDMC gap}~\citep{grosse2015sandwiching}. We note that AIS combined with BDMC has been used to estimate log-likelihoods for deep generative models~\citep{wu2016quantitative}. In this work, we validate our AIS experiments by using the BDMC gap to measure the accuracy of our partition function estimators.

\subsection{Implicit Generative Models}
The goal of generative modeling is to learn a model distribution $p(\mathbf{x})$ to approximate the data distribution $p_d(\mathbf{x})$. Implicit generative models define the model distribution $p(\mathbf{x})$ using a latent variable $\mathbf{z}$ with a fixed prior distribution $p(\mathbf{z})$ such as a Gaussian distribution, and a decoder or generator network which computes $\hat{\mathbf{x}} = f(\mathbf{z})$. In some cases (e.g.~VAEs, AAEs), the generator explicitly parameterizes a conditional distribution $p(\mathbf{x}|\mathbf{z})$, such as a Gaussian observation model $\mathcal{N}(\mathbf{x};f(\mathbf{z}), \sigma^2\mathbf{I})$. But in implicit models such as GANs, the generator directly outputs $\hat{\mathbf{x}} = f(\mathbf{z})$. In order to treat VAEs and GANs under a consistent framework, we ignore the Gaussian observation model of VAEs (thereby treating the VAE decoder as an implicit model), and use the squared error distortion of $d(\mathbf{x},f(\mathbf{z}))=\|\mathbf{x}-f(\mathbf{z})\|_2^2$. However, we note that it is also possible to assume a Gaussian observation model with a fixed $\sigma^2$ for GANs, and use the Gaussian negative log-likelihood (NLL) as the distortion measure for both VAEs and GANs: $d(\mathbf{x},f(\mathbf{z}))=-\log \mathcal{N}(\mathbf{x};f(\mathbf{z}), \sigma^2\mathbf{I})$. This is equivalent to squared error distortion up to a linear transformation.

\subsection{Rate Distortion Theory}
Let $\mathbf{x}$ be a random variable that comes from the data distribution $p_d(\mathbf{x})$. Shannon's fundamental compression theorem states that we can compress this random variable losslessly at the rate of $\mathcal{H}(\mathbf{x})$. But if we allow lossy compression, we can compress $\mathbf{x}$ at the rate of $R$, where $R\leq\mathcal{H}(\mathbf{x})$, using the code $\mathbf{z}$, and have a lossy reconstruction $\hat{\mathbf{x}}=f(\mathbf{z})$ with the distortion of $D$, given a distortion measure $d(\mathbf{x},\hat{\mathbf{x}})=d(\mathbf{x},f(\mathbf{z}))$. The rate distortion theory~\citep{cover2012elements} quantifies the trade-off between the lossy compression rate $R$ and the distortion $D$. The rate distortion function $\mathcal{R}(D)$ is defined as the minimum number of bits per sample required to achieve lossy compression of the data such that the average distortion measured by the distortion function is less than $D$.
Shannon's rate distortion theorem states that $\mathcal{R}(D)$ equals the minimum of the following optimization problem:

\vspace{-.6cm}
\begin{footnotesize}
\begin{align}\label{eq:rd}
\min\limits_{q({\mathbf{z}}|\mathbf{x})} \ \mathcal{I}({\mathbf{z}};\mathbf{x}) \qquad
s.t.\ \mathbb{E}_{q(\mathbf{x},{\mathbf{z}})}[d(\mathbf{x},f(\mathbf{z}))]\leq D.
\end{align}
\end{footnotesize}
\vspace{-.6cm}

where the optimization is over the \emph{channel conditional} distribution $q(\mathbf{z}|\mathbf{x})$. Suppose the data-distribution is $p_d(\mathbf{x})$. The channel conditional $q(\mathbf{z}|\mathbf{x})$ induces the joint distribution $q({\mathbf{z}},\mathbf{x})=p_d(\mathbf{x})q({\mathbf{z}}|\mathbf{x})$, which defines the mutual information $\mathcal{I}({\mathbf{z}};\mathbf{x})$. $q({\mathbf{z}})$ is the marginal distribution over ${\mathbf{z}}$ of the joint distribution $q({\mathbf{z}},\mathbf{x})$, and is called the \emph{output marginal} distribution. 
We can rewrite the optimization of \myeq{rd} using the method of Lagrange multipliers as follows:

\vspace{-.5cm}
\begin{footnotesize}
\begin{align}
\min\limits_{q({\mathbf{z}}|\mathbf{x})} \mathcal{I}({\mathbf{z}};\mathbf{x})
+\beta \mathbb{E}_{q(\mathbf{z},\mathbf{x})}[d(\mathbf{x},f(\mathbf{z}))].\label{eq:rd-lagrange}
\end{align}
\end{footnotesize}
\vspace{-.7cm}

\subsection{Variational Bounds on Mutual Information}
We must modify the standard rate distortion formalism slightly in order to match the goals of generative model evaluation. Specifically, we are interested in evaluating lossy compression with coding schemes corresponding to \emph{particular} trained generative models, including the fixed prior $p(\mathbf{z})$. 
For models such as VAEs, $\mathrm{KL}(q(\mathbf{z}|\mathbf{x})\|p(\mathbf{z}))$ is standardly interpreted as the description length of $\mathbf{z}$. Hence, we adjust the rate distortion formalism to use $\mathbb{E}_{p_d(\mathbf{x})} \mathrm{KL}(q(\mathbf{z}|\mathbf{x})\|p(\mathbf{z}))$ in place of $\mathcal{I}(\mathbf{x}, \mathbf{z})$. It can be shown that $\mathbb{E}_{p_d(\mathbf{x})} \mathrm{KL}(q(\mathbf{z}|\mathbf{x})\|p(\mathbf{z}))$ is a variational upper bound on the standard rate $\mathcal{I}(\mathbf{x}, \mathbf{z})$:

\vspace{-.5cm}
\begin{footnotesize}
\begin{align}\label{eq:rate_upper_bound}\nonumber
\mathcal{I}(\mathbf{x};\mathbf{z}) &\leq \mathcal{I}(\mathbf{x};\mathbf{z}) +\mathrm{KL}( q(\mathbf{z})\|p(\mathbf{z})) \\ 
&=\mathbb{E}_{p_d(\mathbf{x})}\mathrm{KL}( q(\mathbf{z|x})\|p(\mathbf{z})).
\end{align}
\end{footnotesize}
\vspace{-.5cm}

In the context of variational inference, $q(\mathbf{z}|\mathbf{x})$ is the posterior, $q(\mathbf{z}) = \int p_d(\mathbf{x}) q(\mathbf{z}|\mathbf{x}) d\mathbf{x}$ is the aggregated posterior~\citep{makhzani2015adversarial}, and $p(\mathbf{z})$ is the prior.
In the context of rate distortion theory, $q(\mathbf{z}|\mathbf{x})$ is the channel conditional, $q(\mathbf{z})$ is the output marginal, and $p(\mathbf{z})$ is the \emph{variational output marginal} 
distribution.
The inequality is tight when $p(\mathbf{z}) = q(\mathbf{z})$, i.e., the variational output marginal~(prior) is equal to the output marginal~(aggregated posterior). We note that the upper bound of \myeq{rate_upper_bound} has been used in other algorithms such as the Blahut-Arimoto algorithm~\citep{arimoto1972algorithm} or the variational information bottleneck algorithm~\citep{alemi2016deep}.

\begin{figure*}[t]
\begin{minipage}{\linewidth}
    \centering
    \begin{minipage}{0.29\linewidth}
      \begin{figure}[H]
        \begin{center}
        \hspace{-.2cm}\includegraphics[scale=.55]{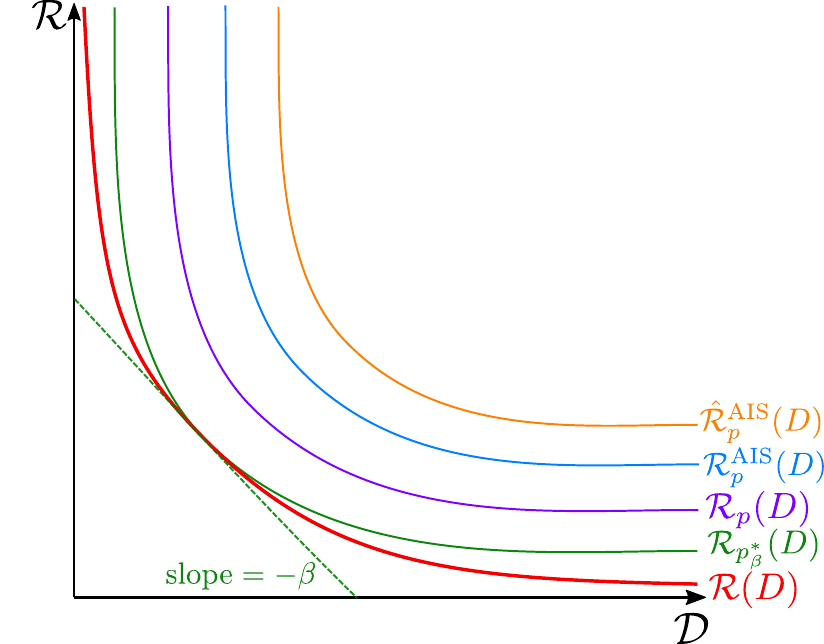}
        \end{center}
        \vspace{-.4cm}
        \caption{\label{fig:rd}\fontsize{9}{1}Geometric illustration of the true RD curve and its upper bounds.}
      \end{figure}
    \end{minipage}
    \hspace{.1cm}
    \begin{minipage}{0.69\linewidth}
      \begin{figure}[H]
      \scriptsize{(a)}\includegraphics[scale=.22]{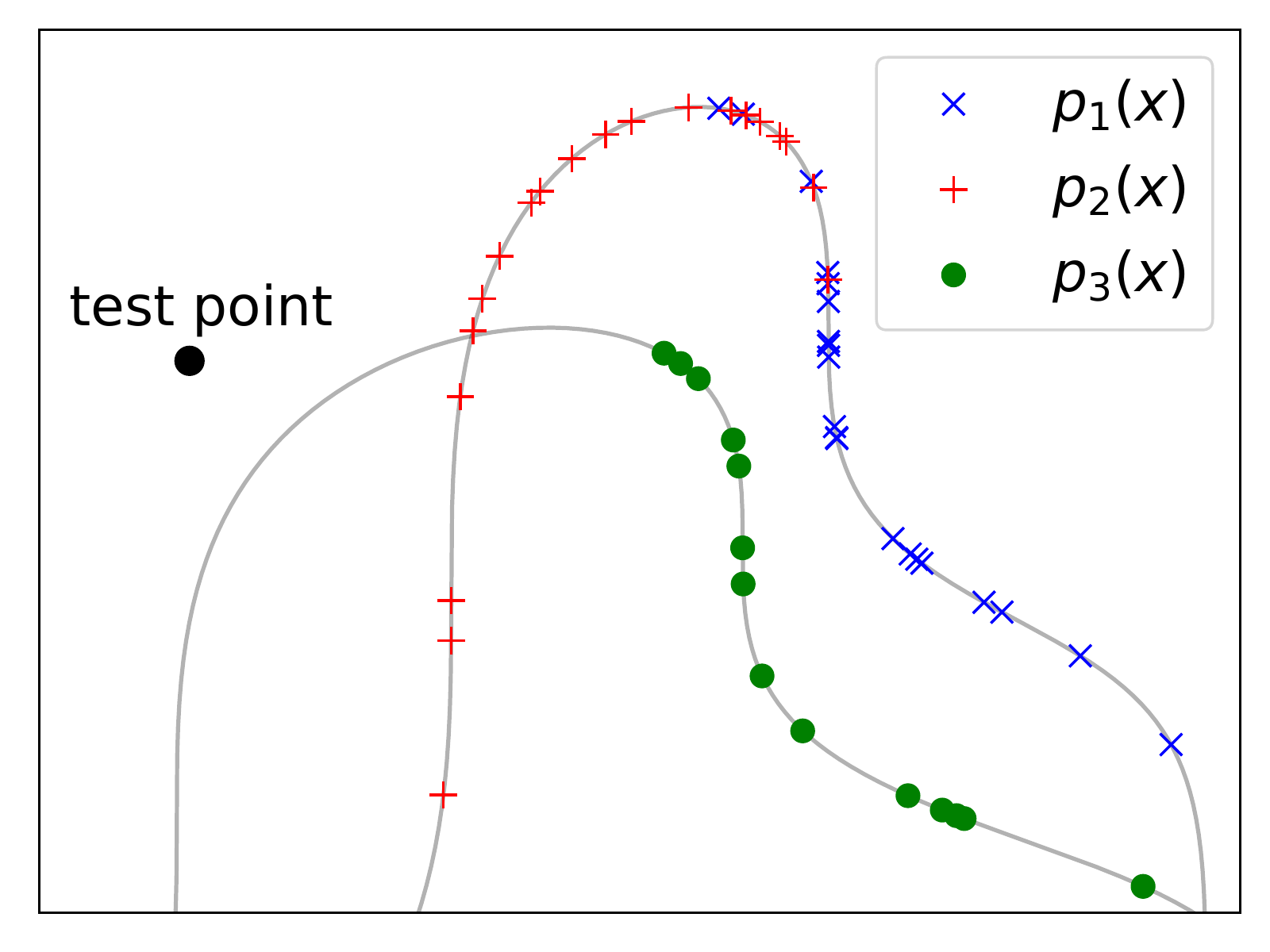}
      \scriptsize{(b)}\includegraphics[scale=.22]{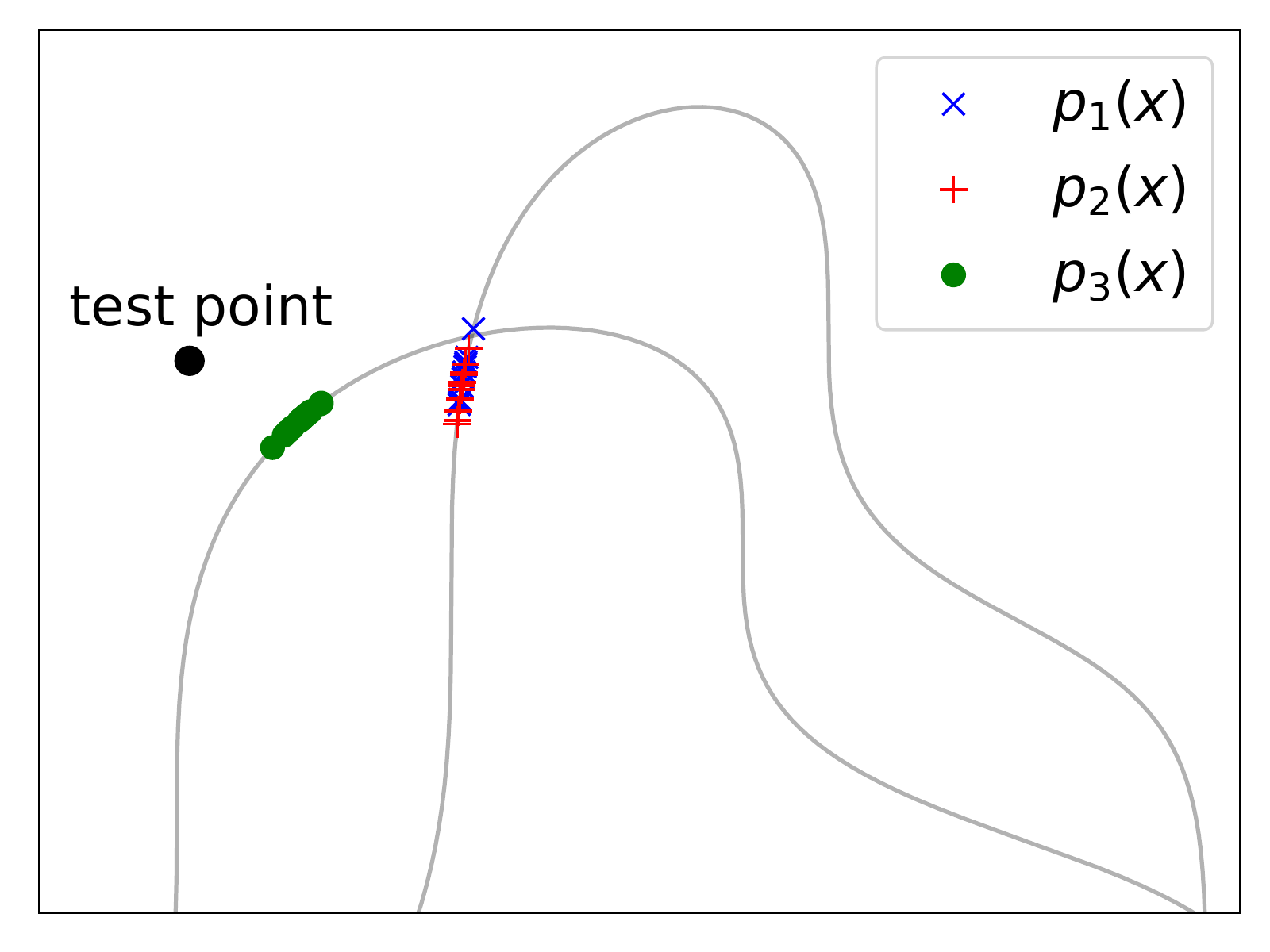}
      \scriptsize{(c)}\includegraphics[scale=.23]{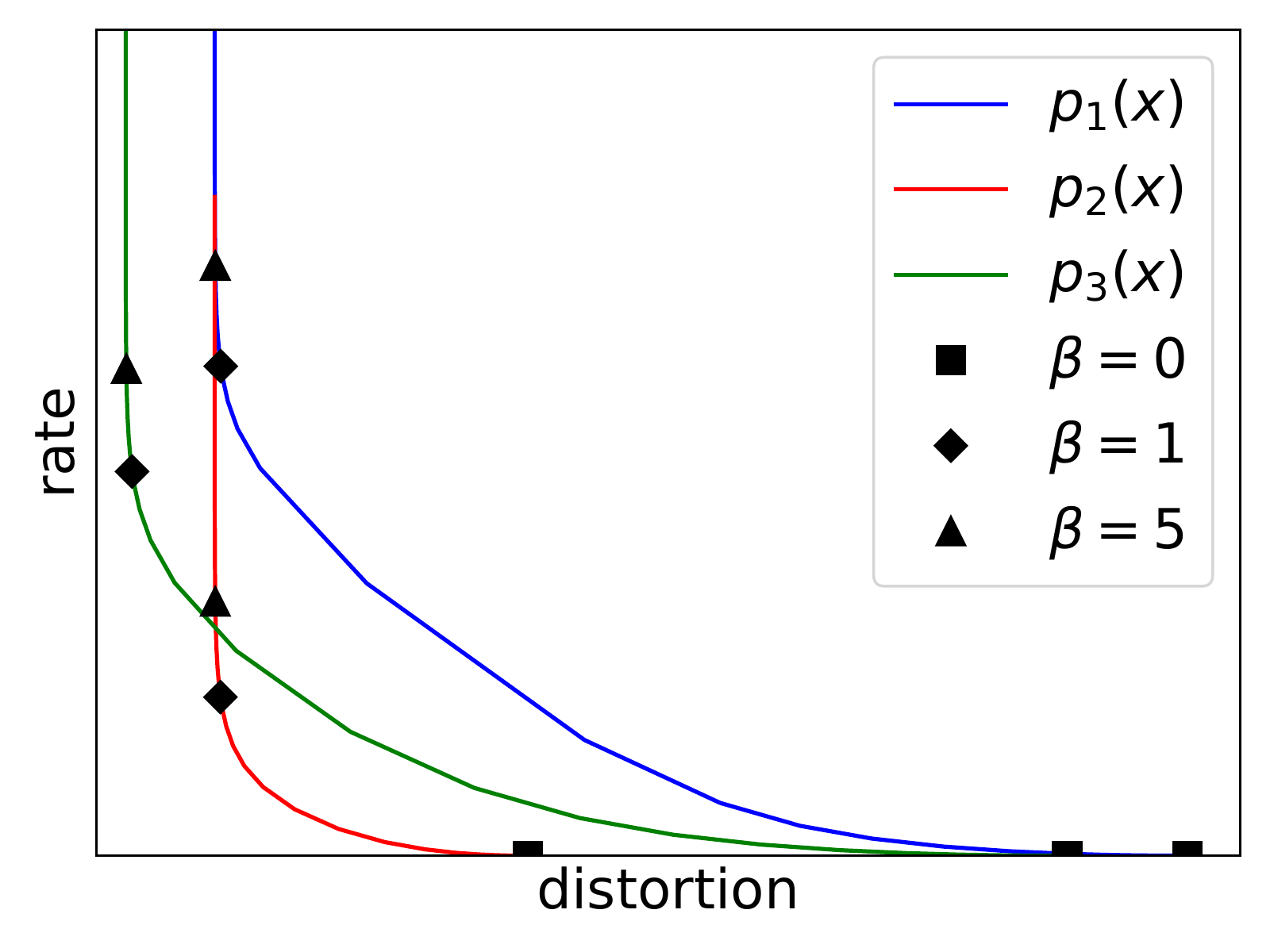}
      \vspace{-.4cm}
      \caption{\label{fig:2d} \textbf{(a)} Prior samples, or equivalently, reconstructions with $\beta=0$. \textbf{(b)} High-rate reconstructions ($\beta=5$). \textbf{(c)} Variational rate distortion curves for each of these three models.}
      \end{figure}
    \end{minipage}
\end{minipage}
\end{figure*}

\section{Variational Rate Distortion Functions}
Analogously to the rate distortion function, we define the \emph{variational rate distortion} function $\mathcal{R}_p(D)$ as the minimum value of the variational upper bound $\mathrm{KL}(q(\mathbf{z}|\mathbf{x})\|p(\mathbf{z}))$ for a given distortion $D$. More precisely, $\mathcal{R}_p(D)$ is the solution of

\vspace{-.5cm}
\begin{footnotesize}
\begin{align}\label{eq:rpd}
\min\limits_{q(\mathbf{z}|\mathbf{x})} &\mathbb{E}_{p_d(\mathbf{x})}\mathrm{KL}( q(\mathbf{z|x})\|p(\mathbf{z})) \quad
s.t.\ \mathbb{E}_{q(\mathbf{x},\mathbf{z})}[d(\mathbf{x},f(\mathbf{z}))]\leq D. 
\end{align}
\end{footnotesize}
We can rewrite the optimization of \myeq{rpd} using the method of Lagrange multipliers as follows:

\vspace{-.5cm}
\begin{footnotesize}
\begin{align}\label{eq:rdl}
\min\limits_{q(\mathbf{z}|\mathbf{x})} \mathbb{E}_{p_d(\mathbf{x})}\mathrm{KL}( q(\mathbf{z|x})\|p(\mathbf{z})) 
+\beta \mathbb{E}_{q(\mathbf{x},\mathbf{z})}[d(\mathbf{x},f(\mathbf{z}))].
\end{align}
\end{footnotesize}
\vspace{-.5cm}

Conveniently, the Lagrangian decomposes into independent optimization problems for each $\mathbf{x}$, allowing us to treat this as an optimization problem over $q(\mathbf{z}|\mathbf{x})$ for fixed $\mathbf{x}$. We can compute the rate distortion curve by sweeping over $\beta$ rather than by sweeping over $D$.

Now we describe some of the properties of the variational rate distortion function $\mathcal{R}_p(D)$, which are straightforward analogues of well-known properties of the rate distortion function.

\parhead{Proposition 1.}\label{prop:1} $\mathcal{R}_p(D)$ has the following properties:
\begin{enumerate}[label=(\alph*)]
\item
$\mathcal{R}_p(D)$ is non-increasing and convex function of~$D$.
\item We have $\mathcal{R}(D) = \min_{p(\mathbf{z})} \mathcal{R}_p(D)$. As a corollary, for any $p(\mathbf{z})$, we have $\mathcal{R}(D) \leq \mathcal{R}_p(D)$.
\item
The variational rate distortion optimization of \myeq{rdl} has a unique global optimum which can be expressed as $q_\beta^{*}(\mathbf{z}|\mathbf{x})={1\over Z_\beta(\mathbf{x})}p(\mathbf{z})\exp(-\beta d(\mathbf{x},f(\mathbf{z})))$.
\end{enumerate}
\emph{Proof.} See \myapp{proof_1}.

\myfig{rd} shows a geometrical illustration of \mypropa{1}. As stated by \mypropa{1}{b}, for any prior $p(\mathbf{z})$, $\mathcal{R}_p(D)$ is a variational upper bound on $\mathcal{R}(D)$. More specifically, we have $\mathcal{R}(D) = \min_{p(\mathbf{z})} \mathcal{R}(D)$, which implies that for any given $\beta$, there exists a prior $p^*_\beta(\mathbf{z})=\int q_\beta^{*}(\mathbf{z}|\mathbf{x}) p_d(\mathbf{x}) d\mathbf{x}$, for which the gap between the rate distortion and variational rate distortion functions at $\beta$ is zero. Furthermore, $\mathcal{R}_{p^*_{\beta}}(D)$ and $\mathcal{R}_p(D)$ are tangent to each other at the point corresponding to $\beta$, and both are tangent to the line with the slope of $-\beta$ passing through this point. In the next section, we will describe how we can use AIS to estimate $\mathcal{R}_p(D)$, and will derive the upper bounds of $\mathcal{R}^{\text{AIS}}_{p}(D)$ and $\hat{\mathcal{R}}^{\text{AIS}}_{p}(D)$.

\parhead{Variational Rate Distortion Functions with NLL Distortion.}
\label{sec:rd_nll}
If the decoder outputs a probability distribution (as in a VAE), we can define the distortion metric to coincide with the negative log-likelihood (NLL): $d(\mathbf{x},f(\mathbf{z})) = -\log p(\mathbf{x}|\mathbf{z})$. We now describe some of the properties of variational rate distortion functions with NLL distortions.

\parhead{Proposition 2.}\label{prop:2} The variational rate distortion function $\mathcal{R}_p(D)$ with NLL distortion of $-\log p(\mathbf{x}|\mathbf{z})$ has the following properties:
\begin{enumerate}[label=(\alph*)]
\item
$\mathcal{R}(D)$ is lower bounded by the linear function of $\mathcal{H}_d - D$, and upper bounded by the variational rate distortion function: $\mathcal{H}_d - D \leq \mathcal{R}(D) \leq \mathcal{R}_p(D).$

\item The global optimum of variational rate distortion optimization (\myeq{rdl}) can be expressed as 

\begin{footnotesize}
$q^{*}_{\beta}(\mathbf{z}|\mathbf{x}) = {1\over Z_\beta(\mathbf{x})}p(\mathbf{z})p(\mathbf{x}|\mathbf{z})^\beta$
\end{footnotesize}

where $Z^{*}_\beta(\mathbf{x}) = \int p(\mathbf{z})p(\mathbf{x}|\mathbf{z})^\beta d\mathbf{z}$.

\item
At $\beta=1$, the negative summation of rate and distortion is the true log-likelihood:

\begin{footnotesize}
$\mathcal{L}_p=\mathbb{E}_{p_d(\mathbf{x})}[\log p(\mathbf{x})]=-R_\beta\big\vert_{\beta=1}-D_\beta\big\vert_{\beta=1}$.
\end{footnotesize}

\end{enumerate}

\emph{Proof.} See \myapp{proof_2}.

\parhead{Illustrative Example.} We now motivate the rate-distortion tradeoff in generative models using an illustrative example (\myfig{2d}). Suppose we have three generative models $p_1(\mathbf{x})$, $p_2(\mathbf{x})$ and $p_3(\mathbf{x})$, on a 2-D space with a 1-D Gaussian latent code. The conditional likelihood of these generative models define a 1-D manifold (grey curves in \myfigab{2d}{a}{b}) on the 2-D data space. In \myfiga{2d}{a}, the colored points represent prior samples from the three models. The $p_1(\mathbf{x})$ and $p_2(\mathbf{x})$ models have the same decoder but have different priors. Thus they define different distributions on the same manifold. The $p_3(\mathbf{x})$ model has a different prior and conditional likelihood. \myfiga{2d}{c} shows the RD curves of each model for the test point shown in \myfigab{2d}{a}{b}. In the low-rate regime, the reconstructions of the models are close to the prior samples (\myfiga{2d}{a}). Thus, the $p_2(\mathbf{x})$ model achieves a better average reconstruction than the $p_3(\mathbf{x})$ model, and the $p_3(\mathbf{x})$ model is better than the $p_1(\mathbf{x})$. However, in the high-rate regime ($\beta=5$, \myfiga{2d}{b}), the $p_3(\mathbf{x})$ model can use a large number of bits to specify reconstruction points on its manifold that are very close to the test-point, and thus outperforms the high-rate reconstructions of both the $p_1(\mathbf{x})$ and $p_2(\mathbf{x})$ model. We can also see that at high-rates, the prior is ignored and the compression rates of the $p_1(\mathbf{x})$ and the $p_2(\mathbf{x})$ model match each other.

\section{\fontsize{11.4}{1}\selectfont Bounding Rate Distortion Functions with AIS}
\label{sec:ais_evaluation}

We've shown that evaluating the variational rate-distortion function $\mathcal{R}_{p}(D)$ amounts to sampling from and estimating the partition functions of a particular sequence of distributions. Unfortunately, both computations are generally intractable for deep generative models. Furthermore, computing the entire RD curve would seem to require separately performing inference for many values of $\beta$.

In this section, we show how to obtain (in practice highly accurate) upper bounds on the RD curve using AIS (see \mysec{ais} for background). AIS has several properties that make it remarkably useful for RD curve estimation: (1) the RD target distributions correspond exactly to the set of intermediate distributions commonly used for AIS, (2) it non-asymptotically lower bounds the log partition function in expectation, and (3) the work already done for one distribution helps in doing inference for the next distribution. Due to these properties, we can obtain the entire RD curve in roughly the same time required to estimate a single log-likelihood value.

\parhead{AIS Chain.}
We fix a temperature schedule $0=\beta_0 < \beta_1 < \ldots < \beta_n=\infty$. For the $k$th intermediate distribution, we use the optimal channel conditional ${q}_k(\mathbf{z}|\mathbf{x})$ and partition function $Z_{k}(\mathbf{x})$, corresponding to points along $\mathcal{R}_{p}(D)$ and derived in \mypropa{1}{c}: 
${q}_k(\mathbf{z}|\mathbf{x}) =\frac{1}{Z_k}\tilde{q}_k(\mathbf{z}|\mathbf{x})$, where

\vspace{-.5cm}
\begin{footnotesize}
\begin{align}
\tilde{q}_k(\mathbf{z}|\mathbf{x})&=p(\mathbf{z})\exp(-\beta_k d(\mathbf{x}, f(\mathbf{z}))), \\
Z_{k}(\mathbf{x})&=\int \tilde{q}_k(\mathbf{z}|\mathbf{x}) d\mathbf{z}.
\end{align}
\end{footnotesize}
\vspace{-.5cm}

Conveniently, this choice coincides with geometric averages, the typical choice of intermediate distributions for AIS.
For the transition operator, we use Hamiltonian Monte Carlo~\citep{neal2011mcmc}.
At the $k$-th step, the rate is denoted by $R_{k}(\mathbf{x})= \mathrm{KL}({q_{k}(\mathbf{z}|\mathbf{x})}\|p(\mathbf{z}))$ and the distortion is denoted by $D_{k}(\mathbf{x})= \mathbb{E}_{q_{k}(\mathbf{z}|\mathbf{x})}[d(\mathbf{x}, f(\mathbf{z}))]$.

\parhead{AIS Rate Distortion Curves.}
For each data point $\mathbf{x}$, we run $M$ independent AIS chains, numbered $i=1, \ldots, M$, in the forward direction. At the $k$-th state of the $i$-th chain, let $\mathbf{z}^i_k$ be the state, ${w}_k^i$ be the AIS importance weights, and $\tilde{w}_k^i$ be the normalized AIS importance weights. We denote the AIS distribution at the $k$-th step as the distribution obtained by first sampling from all the $M$ forward distributions $q_f(\mathbf{z}^i_1, \ldots, \mathbf{z}^i_k|\mathbf{x})\Big\vert_{i=1:M}$, and then re-sampling the samples based on their normalized importance weights $\tilde{w}^{i}_k$ (see \mysec{ais} and \myapp{proof_4} for more details). More formally $q_{k}^{\text{AIS}}(\mathbf{z}|\mathbf{x})$ is

\vspace{-.5cm}
\begin{footnotesize}
\begin{align}
\label{eq:ais_dist}
q_{k}^{\text{AIS}}(\mathbf{z}|\mathbf{x}) = \mathbb{E}_{\prod_{i=1}^M q_f(\mathbf{z}^i_1, \ldots, \mathbf{z}^i_k|\mathbf{x})}[\sum_{i=1}^M \tilde{w}^{i}_k\delta(\mathbf{z}-\mathbf{z}^i_k)].
\end{align}
\end{footnotesize}
\vspace{-.5cm}

Using the AIS distribution $q_{k}^{\text{AIS}}(\mathbf{z}|\mathbf{x})$ defined in \myeq{ais_dist}, we now define the AIS distortion $D_{k}^{\text{AIS}}(\mathbf{x})$ and the AIS rate $R_{k}^{\text{AIS}}(\mathbf{x})$ as follows: $D_{k}^{\text{AIS}}(\mathbf{x}) = \mathbb{E}_{q_{k}^{\text{AIS}}(\mathbf{z}|\mathbf{x})}[d(\mathbf{x}, f(\mathbf{z}))]$ and $R_{k}^{\text{AIS}}(\mathbf{x}) = \mathrm{KL}({q_{k}^{\text{AIS}}(\mathbf{z}|\mathbf{x})}\|p(\mathbf{z}))$.
We now define the \emph{AIS rate distortion curve} $\mathcal{R}^{\text{AIS}}_{p}(D)$ (shown in \myfig{rd}) as the RD curve obtained by tracing pairs of $\big(R_{k}^{\text{AIS}}(\mathbf{x}),D_{k}^{\text{AIS}}(\mathbf{x})\big)$.

\parhead{Proposition 3.}\label{prop:3} The AIS rate distortion curve upper bounds the variational rate distortion function: $\mathcal{R}^{\text{AIS}}_{p}(D) \geq \mathcal{R}_{p}(D)$.
\emph{Proof:} See \myapp{proof_3}.

\parhead{Estimated AIS Rate Distortion Curves.}
Although the AIS distribution can be easily sampled from, its density is intractable to evaluate. As the result, evaluating $\mathcal{R}^{\text{AIS}}_{p}(D)$ is also intractable. We now propose to evaluate an upper-bound on $\mathcal{R}^{\text{AIS}}_{p}(D)$ by finding an upper bound for $R_{k}^{\text{AIS}}(\mathbf{x})$, and an unbiased estimate for $D_{k}^{\text{AIS}}(\mathbf{x})$. We use the AIS distribution samples $\mathbf{z}^i_k$ and their corresponding weights $\tilde{w}_k^i$ to obtain the following distortion and partition function estimates:

\vspace{-.5cm}
\begin{footnotesize}
\begin{align}\nonumber
\hat{D}^{\text{AIS}}_k(\mathbf{x}) = \sum_i \tilde{w}_k^i d(\mathbf{x}, f(\mathbf{z}^i_k))),
\quad
\hat{Z}^{\text{AIS}}_k(\mathbf{x}) = {1 \over M} \sum_i w_k^i.
\end{align}
\end{footnotesize}
\vspace{-.5cm}

Having found the estimates $\hat{D}^{\text{AIS}}_k(\mathbf{x})$ and $\hat{Z}^{\text{AIS}}_k(\mathbf{x})$, we propose to estimate the rate as follows:

\vspace{-.5cm}
\begin{footnotesize}
\begin{align}
\hat{R}^{\text{AIS}}_k(\mathbf{x}) = -\log \hat{Z}^{\text{AIS}}_k(\mathbf{x}) - \beta_k \hat{D}^{\text{AIS}}_k(\mathbf{x}).
\end{align}
\end{footnotesize}
\vspace{-.5cm}

We define the \emph{estimated AIS rate distortion curve} $\hat{\mathcal{R}}^{\text{AIS}}_{p}(D)$ (shown in \myfig{rd}) as an RD curve obtained by tracing pairs of rate distortion estimates $\big(\hat{R}_{k}^{\text{AIS}}(\mathbf{x}),\hat{D}_{k}^{\text{AIS}}(\mathbf{x})\big)$.

\parhead{Proposition 4.}\label{prop:4} The estimated AIS rate distortion curve upper bounds the AIS rate distortion curve in expectation: $\mathbb{E}[\hat{\mathcal{R}}^{\text{AIS}}_{p}(D)] \geq \mathcal{R}^{\text{AIS}}_{p}(D)$.  More specifically, we have

\vspace{-.5cm}
\begin{footnotesize}
\begin{align}
\mathbb{E}[\hat{R}_{k}^{\text{AIS}}(\mathbf{x})]\geq R_{k}^{\text{AIS}}(\mathbf{x}),
\qquad
\mathbb{E}[\hat{D}_{k}^{\text{AIS}}(\mathbf{x})] = D_{k}^{\text{AIS}}(\mathbf{x}).
\end{align}
\end{footnotesize}
\vspace{-.5cm}

\emph{Proof Sketch.} For the complete proof see \myapp{proof_4}. It is straightforward to show that $\mathbb{E}[\hat{D}_{k}^{\text{AIS}}(\mathbf{x})] = D_{k}^{\text{AIS}}(\mathbf{x})$, however, the nontrivial part is to show $\mathbb{E}[\hat{R}_{k}^{\text{AIS}}(\mathbf{x})]\geq R_{k}^{\text{AIS}}(\mathbf{x})$. Suppose $\mathbf{V}$ is all the AIS states across $M$ parallel chains except the final selected state $\mathbf{z}$. The monotonicity of KL divergence enables us to upper bound the rate $R_{k}^{\text{AIS}}(\mathbf{x}) = \mathrm{KL}({q_{k}^{\text{AIS}}(\mathbf{z}|\mathbf{x})}\|p(\mathbf{z}))$ in terms of the full KL divergence in the AIS extended state space: $\mathrm{KL}(q_{k}^{\text{AIS}}(\mathbf{z}, \mathbf{V}|\mathbf{x}) \| p(\mathbf{z}) p(\mathbf{V} | \mathbf{z}, \mathbf{x}))$, where $q_{k}^{\text{AIS}}(\mathbf{z}, \mathbf{V}|\mathbf{x})$ is constructed to have the marginal of $q_{k}^{\text{AIS}}(\mathbf{z}|\mathbf{x})$ (similar to the auxiliary variable construction of \citet{domke2018importance}), and $p(\mathbf{z}) p(\mathbf{V} | \mathbf{z}, \mathbf{x})$ has the marginal of $p(\mathbf{z})$. We then prove that the AIS estimate of the rate $\hat{R}_{k}^{\text{AIS}}(\mathbf{x})$ equals the full KL divergence in expectation (see \myapp{proof_4}).

In summary, from \myprop{1}, \myprop{3} and \myprop{4}, we can conclude that the estimated AIS rate distortion curve upper bounds the true rate distortion curve in expectation (shown in \myfig{rd}):
\begin{align}
\mathbb{E}[\hat{\mathcal{R}}^{\text{AIS}}_{p}(D)] \geq \mathcal{R}^{\text{AIS}}_{p}(D) \geq \mathcal{R}_p(D) \geq \mathcal{R}(D).
\end{align}
In all our experiments, we plot the estimated AIS rate distortion function $\hat{\mathcal{R}}^{\text{AIS}}_{p}(D)$.

\parhead{Accuracy of AIS Estimates.} While the above discussion focuses on obtaining upper bounds, we note that AIS is one of the most accurate general-purpose methods for estimating partition functions, and therefore we believe our AIS upper bounds to be fairly tight in practice. In theory, for a large number of intermediate distributions, the AIS variance is proportional to $1/MK$~\citep{neal2001annealed,neal2005estimating}, where $M$ is the number of AIS chains and $K$ is the number of intermediate distributions. For the main experiments of our paper, we evaluate the tightness of the AIS estimate by computing the BDMC gap, and show that in practice our upper bounds are tight (\mysec{validate} and \myapp{experiment_ais}).

\begin{figure*}[t]
\begin{minipage}{\linewidth}
    \centering
    \begin{minipage}{0.30\linewidth}
      \begin{figure}[H]
        \begin{center}
          \includegraphics[scale=.33]{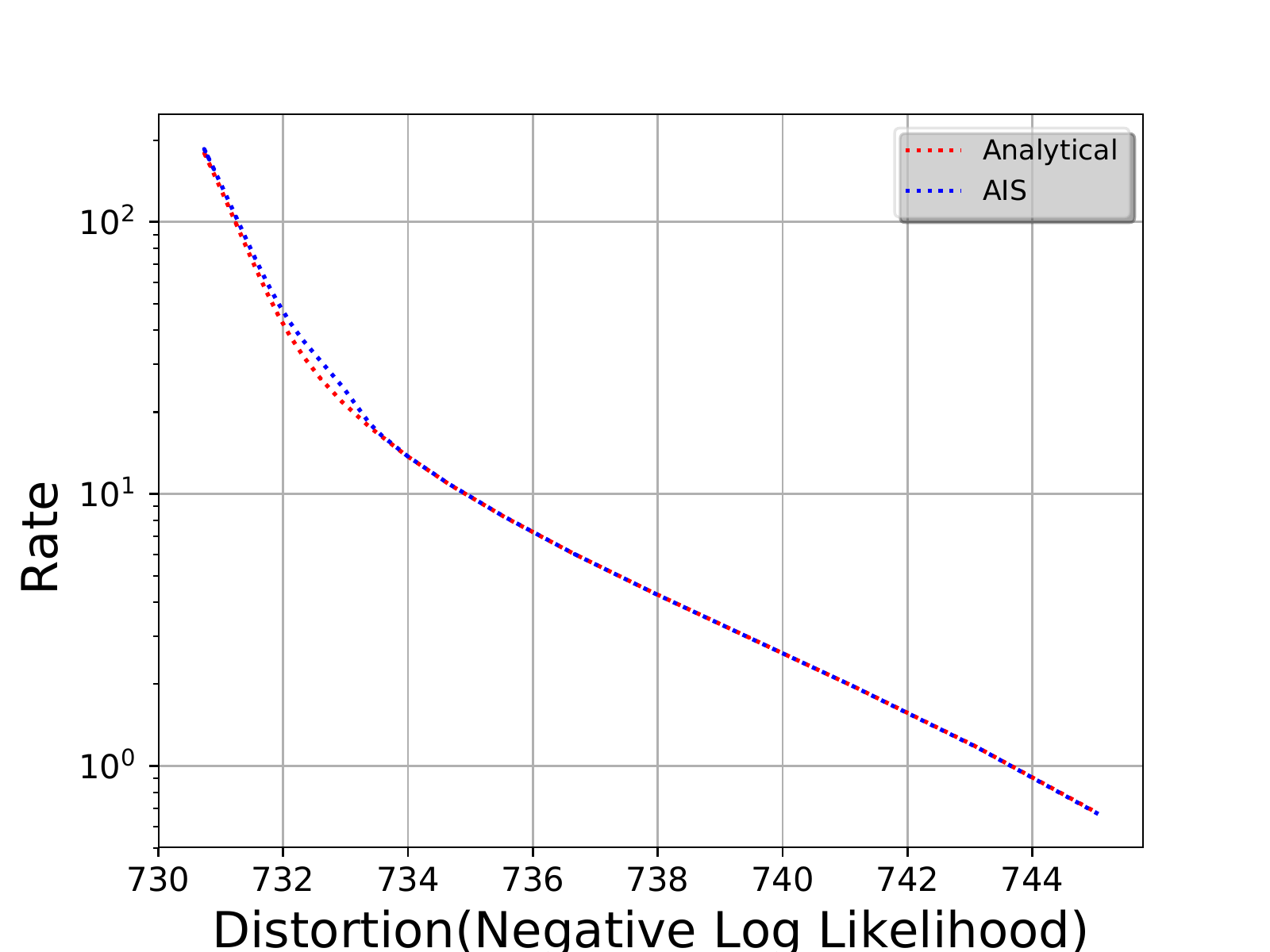}
        \end{center}
        \vspace{-.3cm}
        \caption{\label{fig:vae_analytical}Analytical vs.~AIS variational RD curves for a linear VAE.}
      \end{figure}    
    \end{minipage}
    \hspace{.1cm}
    \begin{minipage}{0.68\linewidth}
      \begin{figure}[H]
      \centering
      \subfigure[]{
      \includegraphics[scale=.35]{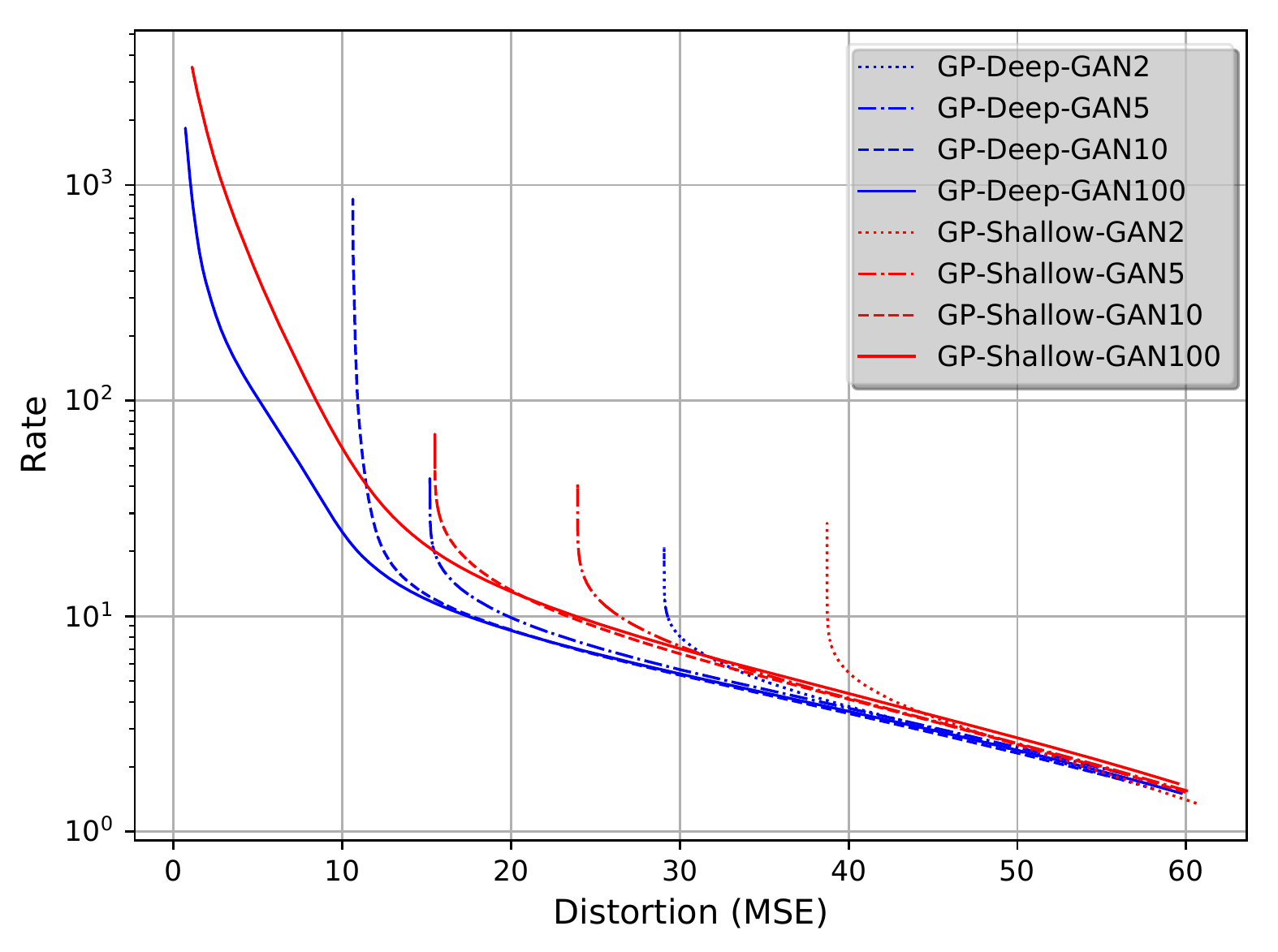}}
      \subfigure[]{
      \includegraphics[scale=.35]{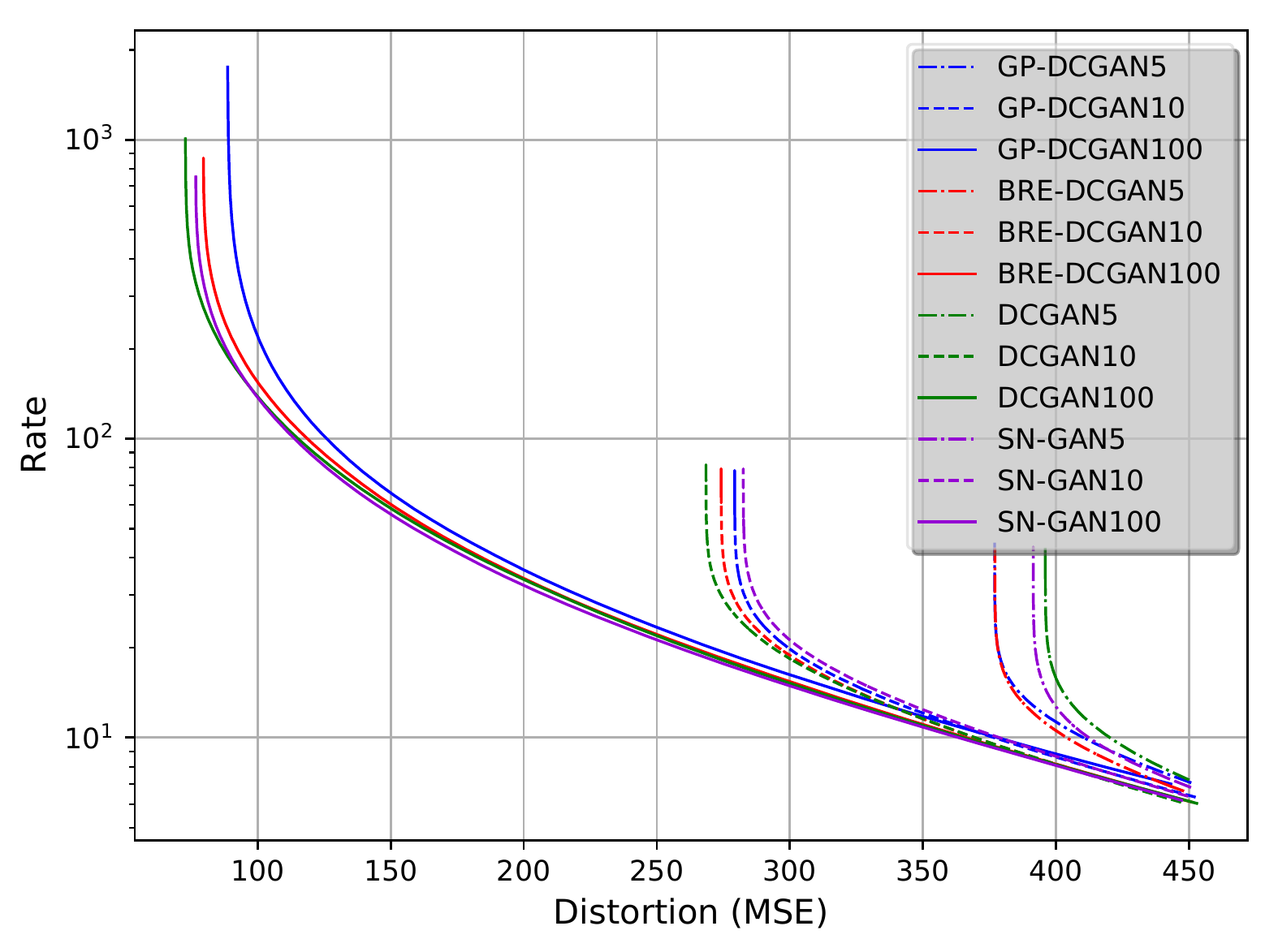}}
      \vspace{-.4cm}
      \caption{\label{fig:gan_mnist_cifar}Variational rate-distortion curves of GANs. \textbf{(a)} MNIST. \textbf{(b)} CIFAR-10.}
      \end{figure}
    \end{minipage}
\end{minipage}
\end{figure*}

\begin{figure}
  \begin{center}
  \hspace{-.2cm}\includegraphics[scale=.43]{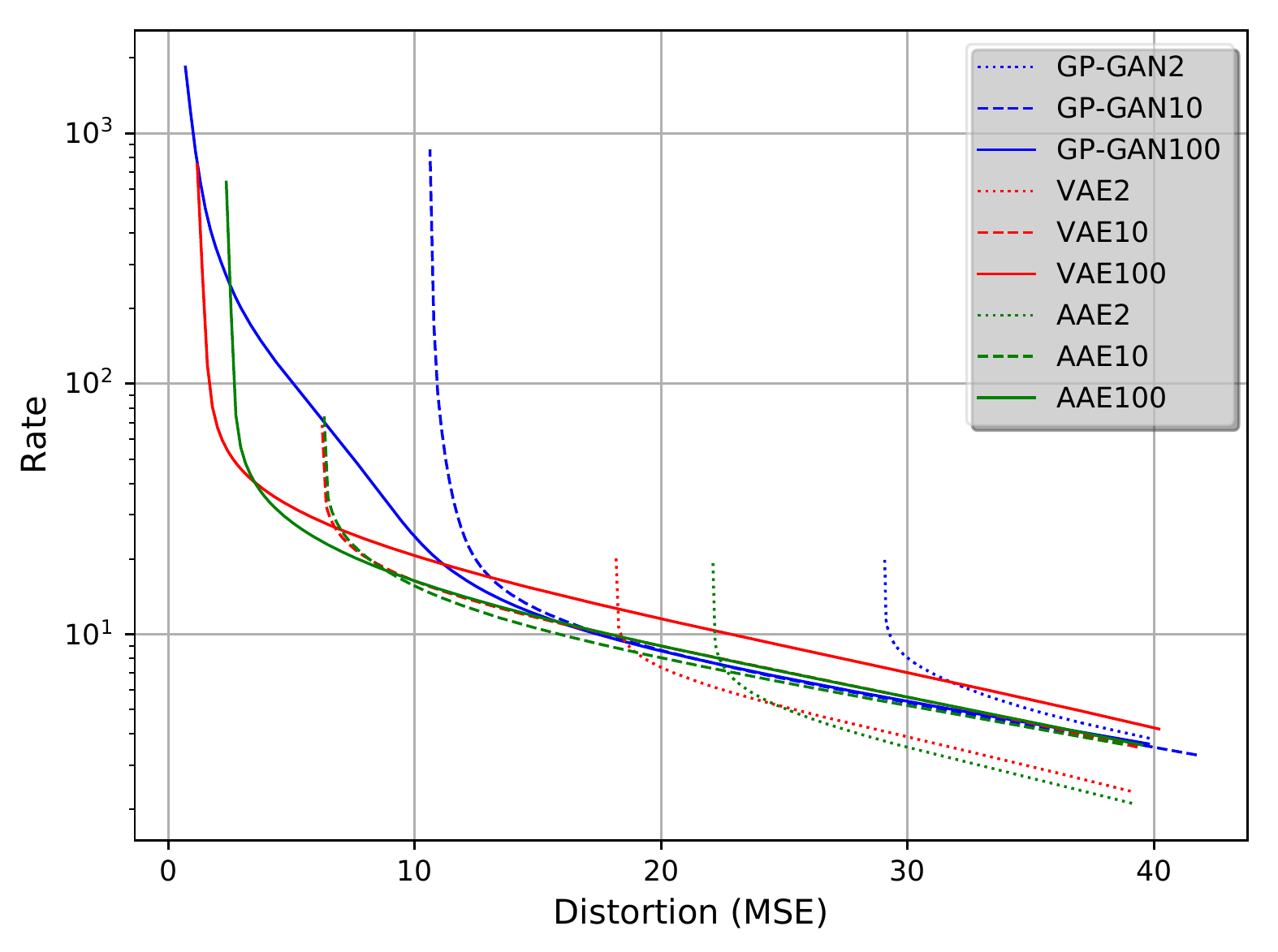}
  \end{center}
  \vspace{-.4cm}
  \caption{\label{fig:vae_gan_mnist}\fontsize{9}{1}Variational RD curves of VAEs, GANs and AAEs.}
\end{figure}

\vspace{-.3cm}
\section{Related Work}
\label{sec:related_works}
\vspace{-.1cm}

\parhead{Evaluation of Implicit Generative Models.} Many heuristic measures have been proposed for evaluation of implicit models, such as the Inception score~\citep{salimans2016improved} and the Fr\'{e}chet Inception Distance (FID)~\citep{fid}.
One of the main drawbacks of the IS or FID is that they can only provide a single scalar value that cannot distinguish the mode dropping behavior from the mode inventing behavior in generative models. In order to address this, \citet{sajjadi2018assessing} proposed to study the precision-recall tradeoff for evaluating generative models. The precision-recall tradeoff is analogous to our rate-distortion tradeoff, but has a very different mathematical motivation.

\parhead{Rate Distortion Theory and Generative Models.} Perhaps the closest work to ours is ``Fixing a Broken ELBO''~\citep{alemi2018fixing}, which plots variational rate distortion curves for VAEs. Our work is different than~\citet{alemi2018fixing} in two key aspects. First, in~\citet{alemi2018fixing} the variational rate distortion function is evaluated by fixing the architecture of the neural network, and learning the distortion measure $d(\mathbf{x},f(\mathbf{z}))$ in addition to learning $q(\mathbf{z}|\mathbf{x})$. Whereas, our work follows a conceptually different goal which is to evaluate a particular generative model with a fixed prior and decoder, independent of how the model was trained. The second key difference is that we find the optimal channel conditional $q^{*}(\mathbf{z}|\mathbf{x})$ by using AIS; while in~\citet{alemi2018fixing}, $q(\mathbf{z}|\mathbf{x})$ is a variational distribution restricted to a variational family. In \mysec{var}, we will empirically show that AIS obtains significantly tighter RD bounds than variational methods. 

\parhead{Practical Compression Schemes.} We have justified our use of compression terminology in terms of Shannon's fundamental result implying that there exists a rate distortion code for any rate distortion pair that is achievable according to the rate distortion function. For \emph{lossless} compression with generative models, there is a practical compression scheme which nearly achieves the theoretical rate (i.e.~the negative ELBO): bits-back encoding. The basic scheme was proposed by~\citet{wallace1990classification, hinton1993keeping}, and later implemented by~\citet{frey1996free}. Practical versions for modern deep generative models were developed by~\citet{townsend2019practical,kingma2019bit}. Other researchers have developed practical \emph{lossy} coding schemes achieving variational rate distortion bounds for particular latent variable models which exploited the factorial structure of the variational posterior~\citep{balle2018variational, theis2017lossy, yang2020improving}. These methods are not directly applicable in our setting, since we don't assume an explicit encoder network, and our variational posteriors lack a convenient factorized form. We don't know whether our variational approximation will lead to a practical lossy compression scheme, but the successes for other variational methods give us hope.

\vspace{-.3cm}
\section{Experiments}
\label{sec:experiment}
\vspace{-.1cm}

In this section, we use our rate distortion approximations to answer the following questions: How do different generative models such as VAEs, GANs and AAEs perform at different lossy compression rates? What insights can we obtain from the rate distortion curves about different characteristics of generative models? What is the effect of the code size (width), depth of the network, or the learning algorithm on the rate distortion tradeoffs? 

The code for reproducing the experiments can be found at \href{https://github.com/BorealisAI/rate_distortion}{https://github.com/BorealisAI/rate\_distortion} and \href{https://github.com/huangsicong/rate_distortion}{https://github.com/huangsicong/rate\_distortion}.

\vspace{-.2cm}
\subsection{Validating AIS}
\label{sec:validate}
\vspace{-.1cm}
\parhead{Linear VAE and BDMC.} We conducted several experiments to validate the correctness of our implementation and the accuracy of the AIS estimates. Firstly, we compared our AIS results with the analytical solution of the variational rate distortion curve on a linear VAE (derived in \myapp{analytical}) trained on MNIST. As shown in \myfig{vae_analytical}, the RD curve estimated by AIS agrees closely with the analytical solution. Secondly, for the main experiments of the paper, we evaluated the tightness of the AIS estimate by computing the BDMC gap. The largest BDMC gap for VAEs and AAEs was 0.537 nats, and the largest BDMC gap for GANs was 3.724 nats, showing that our AIS upper bounds are tight. More details are provided in \myapp{experiment_ais}.

\vspace{-.2cm}
\subsection{Rate Distortion Curves of Deep Generative Models}
\vspace{-.1cm}
\parhead{Rate Distortion Curves of GANs.} \myfig{gan_mnist_cifar} shows rate distortion curves for GANs trained on MNIST and CIFAR-10. We varied the dimension of the noise vector $\mathbf{z}$, as well as the depth of the decoder. For the GAN experiments on MNIST (\myfiga{gan_mnist_cifar}{a}), the label ``deep'' corresponds to three hidden layers of size 1024, and the label ``shallow'' corresponds to one hidden layer of size 1024. We trained shallow and deep GANs with Gradient Penalty (GAN-GP)~\citep{gulrajani2017improved} with the code size $d \in \{2,5,10,100\}$ on MNIST. For the GAN experiments on CIFAR-10 (\myfiga{gan_mnist_cifar}{b}), we trained the DCGAN~\citep{radford2015unsupervised}, GAN with Gradient Penalty (GP)~\citep{gulrajani2017improved}, SN-GAN~\citep{miyato2018spectral}, and BRE-GAN~\citep{BRE2018}, with the code size of $d \in \{2,10,100\}$. In both the MNIST and CIFAR experiments, we observe that in general increasing the code size has the effect of \emph{extending the curve leftwards}. 
This is expected, since the high-rate regime is effectively measuring reconstruction ability, and additional dimensions in $\mathbf{z}$  improves the reconstruction. 

We also observe from \myfiga{gan_mnist_cifar}{b} that different GAN variants with the same code size have nearly identical RD curves, and that the code size dominates the algorithmic differences of GANs.

We can also observe from \myfiga{gan_mnist_cifar}{a} that increasing the depth pushes the curves \emph{down and to the left}. In other words, the distortion in both high-rate and mid-rate regimes improves. In these regimes, increasing the depth increases the capacity of the network, which enables the network to make a better use of the information in the code space. In the low-rate regime, however, increasing the depth, similar to increasing the latent size, does not improve the distortion.

\parhead{Rate Distortion Curves of VAEs.} \myfig{vae_gan_mnist} compares VAEs, AAEs and GP-GANs~\citep{gulrajani2017improved} with the code size of $d \in \{2,10,100\}$, and the same decoder architecture on the MNIST dataset. 
In general, we can see that in the mid-rate to high-rate regimes, VAEs achieve better distortions than GANs with the same architecture. This is expected as the VAE is trained with the ELBO objective, which encourages good reconstructions (in the case of factorized Gaussian decoder).
We can see from \myfig{vae_gan_mnist} that in VAEs, increasing the latent capacity pushes the rate distortion curve \emph{up and to the left}. In other words, in contrast with GANs where increasing the latent capacity always improves the rate distortion curve, in VAEs, there is a trade-off whereby increasing the capacity reduces the distortion at the high-rate regime, at the expense of increasing the distortion in the low-rate regime (or equivalently, increasing the rate required to adequately approximate the data).

We believe the performance drop of VAEs in the low-rate regime is symptomatic of the ``holes problem''~\citep{rezende2018taming, makhzani2015adversarial} in the code space of VAEs with large code size: because these VAEs allocate a large fraction of their latent spaces to garbage images, it requires many bits to get close to the image manifold. Interestingly, this trade-off could also help explain the well-known problem of blurry samples from VAEs: in order to avoid garbage samples (corresponding to large distortion in the low-rate regime), one needs to reduce the capacity, thereby increasing the distortion at the high-rate regime. By contrast, GANs do not suffer from this tradeoff, and one can train high-capacity GANs without sacrificing performance in the low-rate regime.

\begin{figure*}[t]
\centering
(a)\includegraphics[scale=.35]{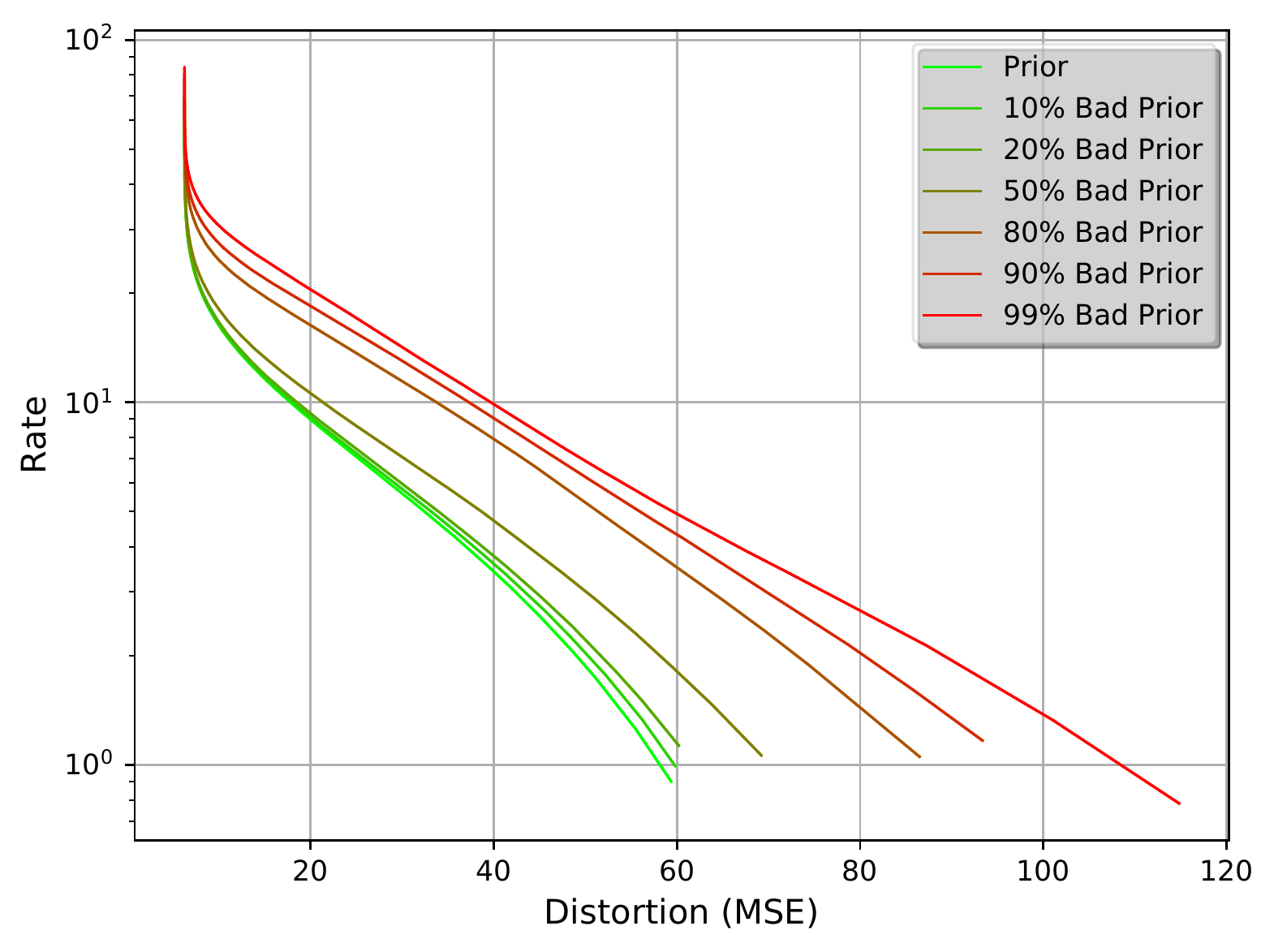}
\hspace{2cm}
(b)\includegraphics[scale=.35]{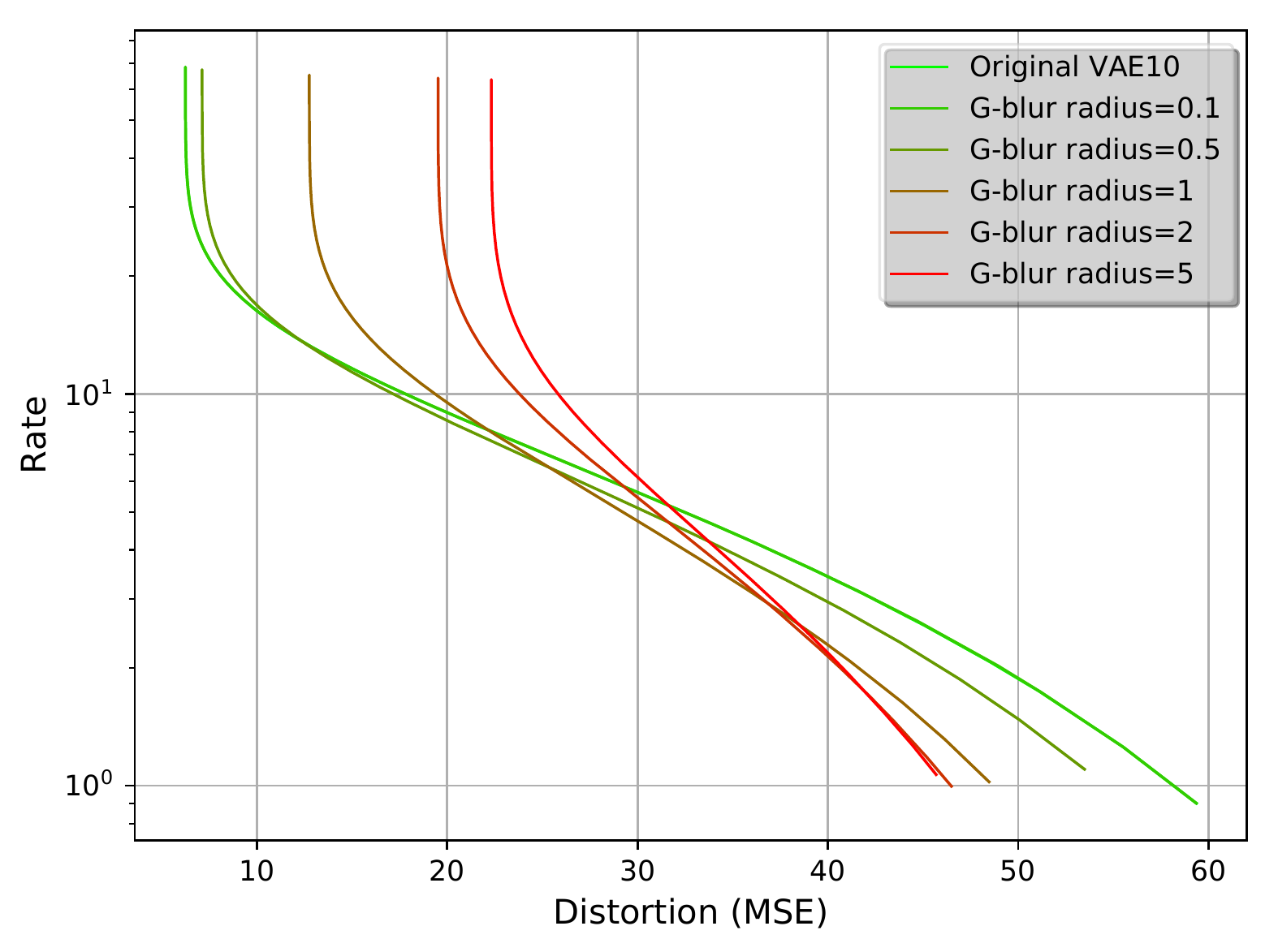}
\vspace{-.3cm}
\caption{\label{fig:mixture_blur}\textbf{(a)} Effect of damaging the VAE prior by using a mixture with a bad prior. \textbf{(b)} Effect of damaging the conditional likelihood of a VAE by convolving with a Gaussian blur kernel after the last decoder layer.}
\end{figure*}

\parhead{Rate Distortion Curves of AAEs.} The AAE was introduced by~\citet{makhzani2015adversarial} to address the holes problem of VAEs, by directly matching the aggregated posterior to the prior in addition to optimizing the reconstruction cost. \myfig{vae_gan_mnist} shows the RD curves of AAEs. In comparison to GANs, AAEs can match the low-rate performance of GANs, but achieve a better high-rate performance. This is expected as AAEs directly optimize the reconstruction cost as part of their objective. In comparison to VAEs, AAEs perform slightly worse at the high-rate regime, which is expected as the adversarial regularization of AAEs is stronger than the KL regularization of VAEs. But AAEs perform slightly better in the low-rate regime, as they can alleviate the holes problem to some extent.

\vspace{-.3cm}
\subsection{Distinguishing Different Failure Modes in Generative Modeling}
\vspace{-.1cm}
Since log-likelihoods constitute only a scalar value, they are unable to distinguish different aspects of a generative model which could be good or bad, such as the prior or the observation model. Here, we show that two manipulations which damage a trained VAE in different ways result in very different behavior of the RD curves.

Our first manipulation, originally proposed by~\citet{theis2015note}, is to use a mixture of the VAE's density and another distribution concentrated away from the data distribution. As pointed out by~\citet{theis2015note}, this results in a model which achieves high log-likelihood while generating poor samples.
Specifically, after training the VAE10 on MNIST, we ``damage'' its prior $p(\mathbf{z})=\mathcal{N}(0,\mathbf{I})$ by altering it to a mixture prior $(1-\alpha) p(\mathbf{z})+\alpha q(\mathbf{z})$, where $q(\mathbf{z})=\mathcal{N}(0,10\mathbf{I})$ is a ``poor'' prior, which is chosen to be far away from the original prior $p(\mathbf{z})$; and $\alpha$ is close to 1. This process would results in a ``poor'' generative model that generates garbage samples most of the time (more precisely with the probability of $\alpha$). Suppose $p(\mathbf{x})$ and $q(\mathbf{x})$ are the likelihood of the good and the poor generative models. It is straightforward to see that $\log q(\mathbf{x})$ is at most $4.6$ nats worse that $\log p(\mathbf{x})$, and thus log-likelihood fails to tell these models apart:

\vspace{-.7cm}
\begin{footnotesize}
\begin{align}\label{eq:bad_model}
\log q(\mathbf{x}) &= \log \Big(0.01 p(\mathbf{x})+0.99 \int q(\mathbf{z}) p(\mathbf{x}|\mathbf{z}) d\mathbf{z}\Big)\\
&>\log (0.01 p(\mathbf{x})) \approx \log p(\mathbf{x}) - 4.6
\end{align}
\end{footnotesize}
\vspace{-.7cm}

\myfiga{mixture_blur}{a} plots the RD curves of this model for different values of $\alpha$. We can see that the high-rate and log-likelihood performance of the good and poor generative models are almost identical, whereas in the low-rate regime, the RD curves show a significant drop in the performance and thus successfully detect this failure mode of log-likelihood.

Our second manipulation is to damage the decoder by convolving its output with a Gaussian blur kernel. \myfiga{mixture_blur}{b} shows the rate distortion curves for different radii of the Gaussian kernel. We can see that, in contrast to the mixture prior experiment, the high-rate performance of the VAE drops due to inability of the decoder to output sharp images. However, we can also see an ``improvement'' in the low-rate performance of the VAE. This is because the data distribution does not necessarily achieve the minimal distortion, and in fact, in the extremely low-rate regime, blurring appears to help by reducing the average Euclidean distance between low-rate reconstructions and the input images. This problem was observed by \citet{theis2015note} in the context of log-likelihood estimation; our current observation indicates that rate distortion analysis does not fix this problem.

Our two manipulations --- bad priors and blurring --- resulted in very different changes to the RD curves, suggesting that these curves provide a richer picture of the performance of generative models, compared to scalar metrics such as log-likelihoods or FID.

\begin{figure*}[t]
\centering
(a)\includegraphics[scale=.35]{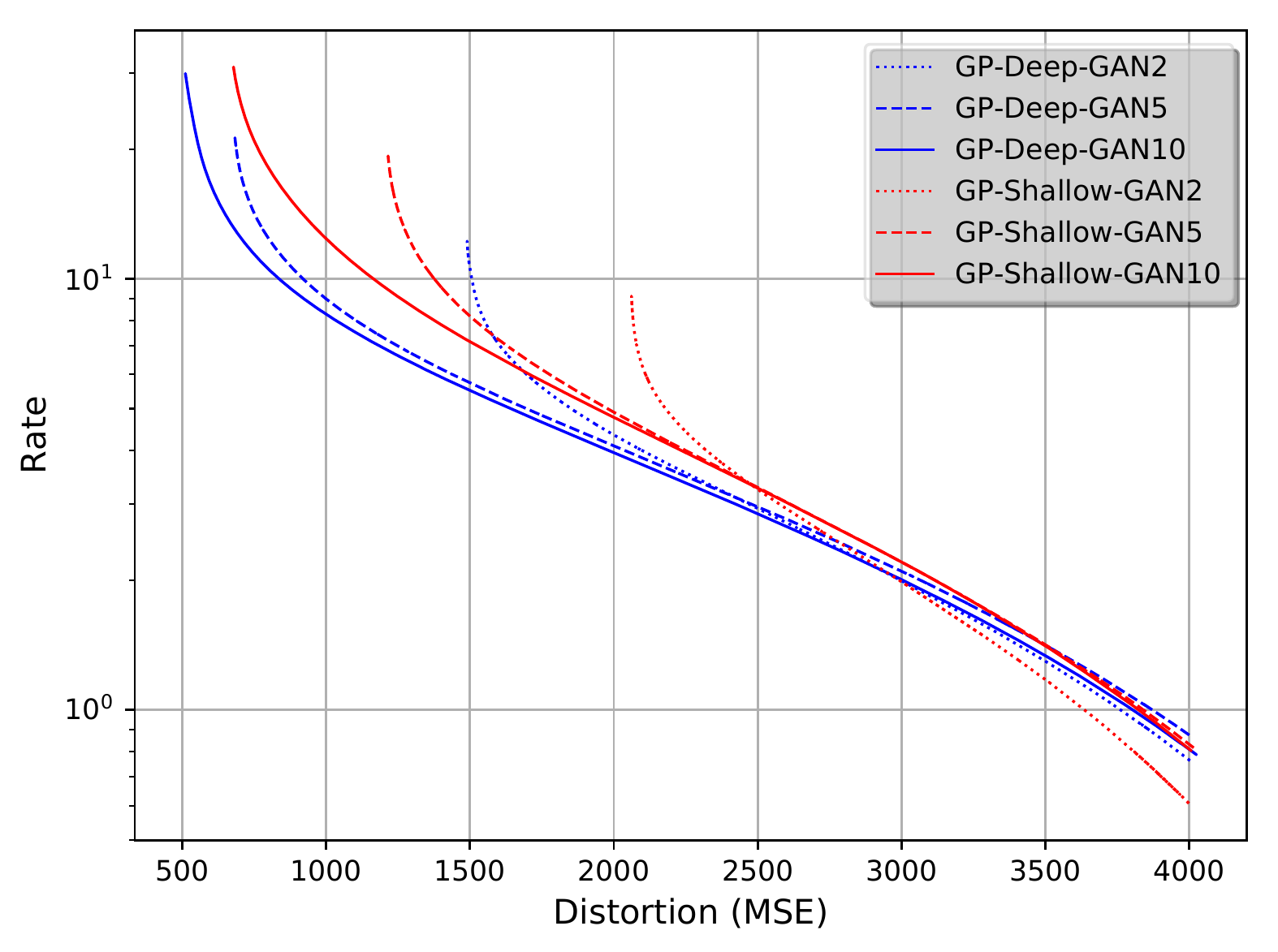}
\hspace{2cm}
(b)\includegraphics[scale=.35]{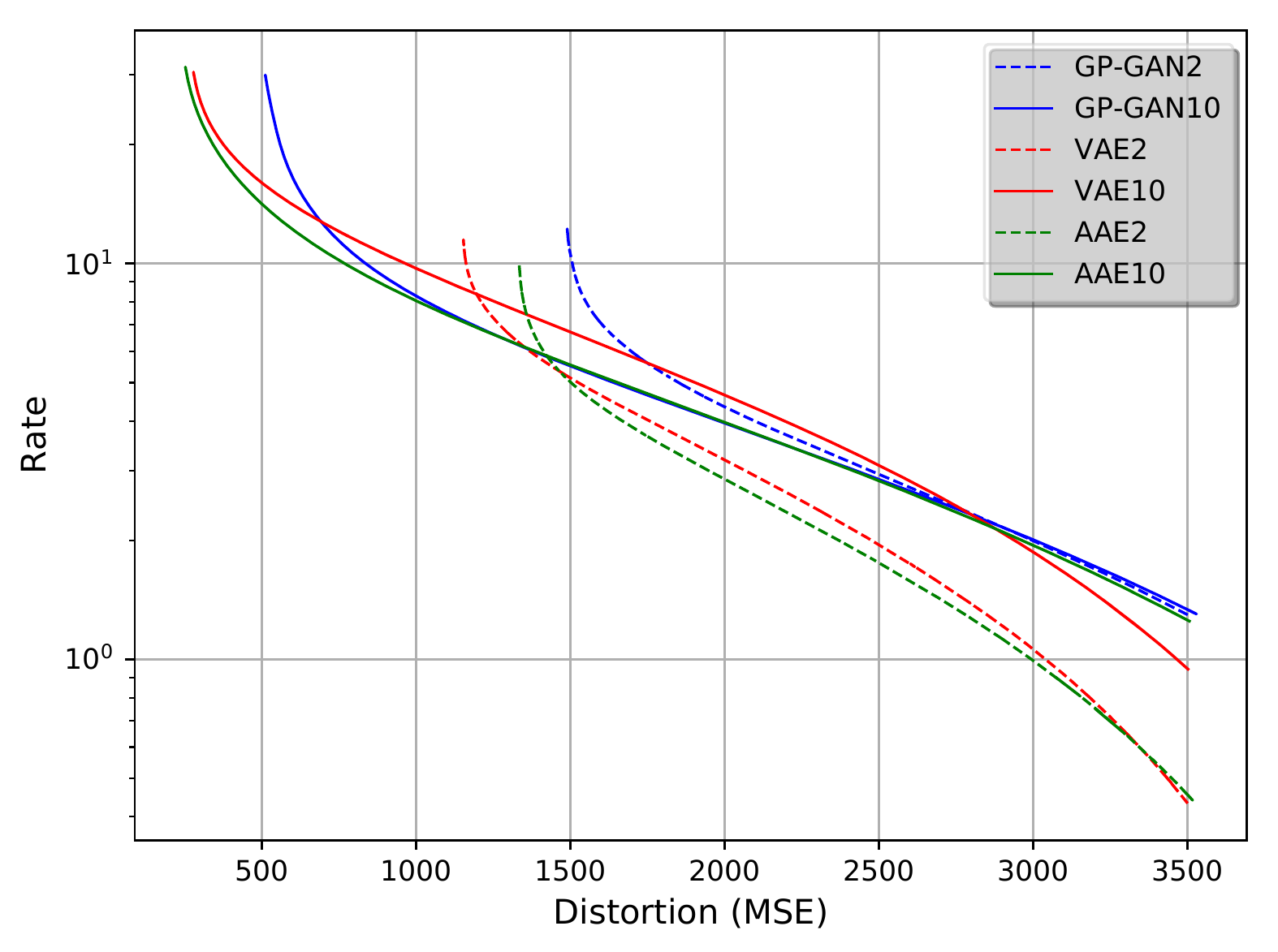}
\vspace{-.3cm}
\caption{\label{fig:deep_feature_mnist}The RD curves of GANs, VAEs and AAEs with MSE distortion on the deep feature space. The behavior is qualitatively similar to the results for MSE in images (see \myfig{gan_mnist_cifar} and  \myfig{vae_gan_mnist}), suggesting that the RD analysis is not particularly sensitive to the particular choice of metric. \textbf{(a)} GANs. \textbf{(b)} VAEs, GANs, and AAEs.}
\end{figure*}

\begin{figure*}[t]
\centering
\centering
(a)\includegraphics[scale=.35]{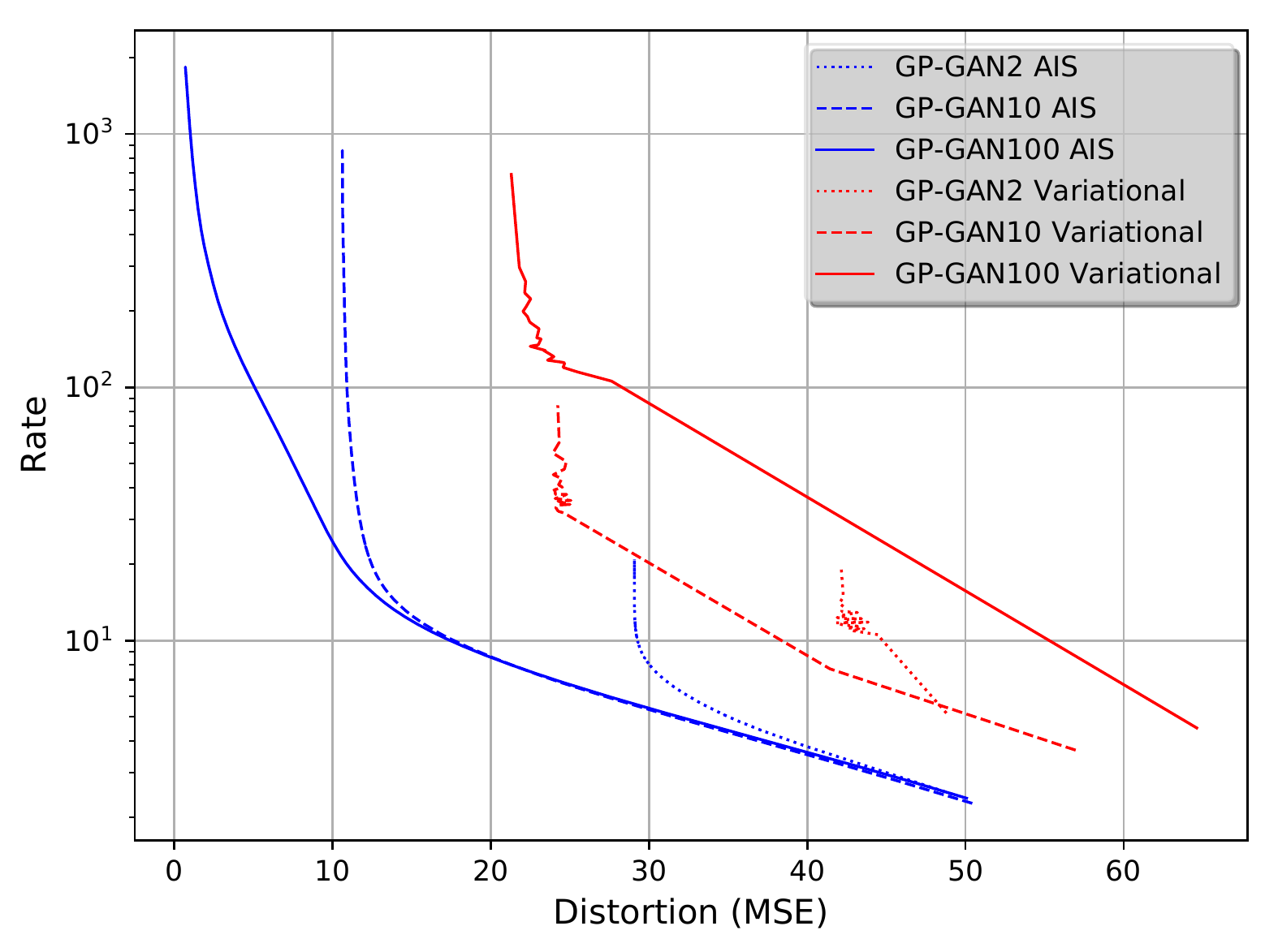}
\hspace{2cm}
(b)\includegraphics[scale=.35]{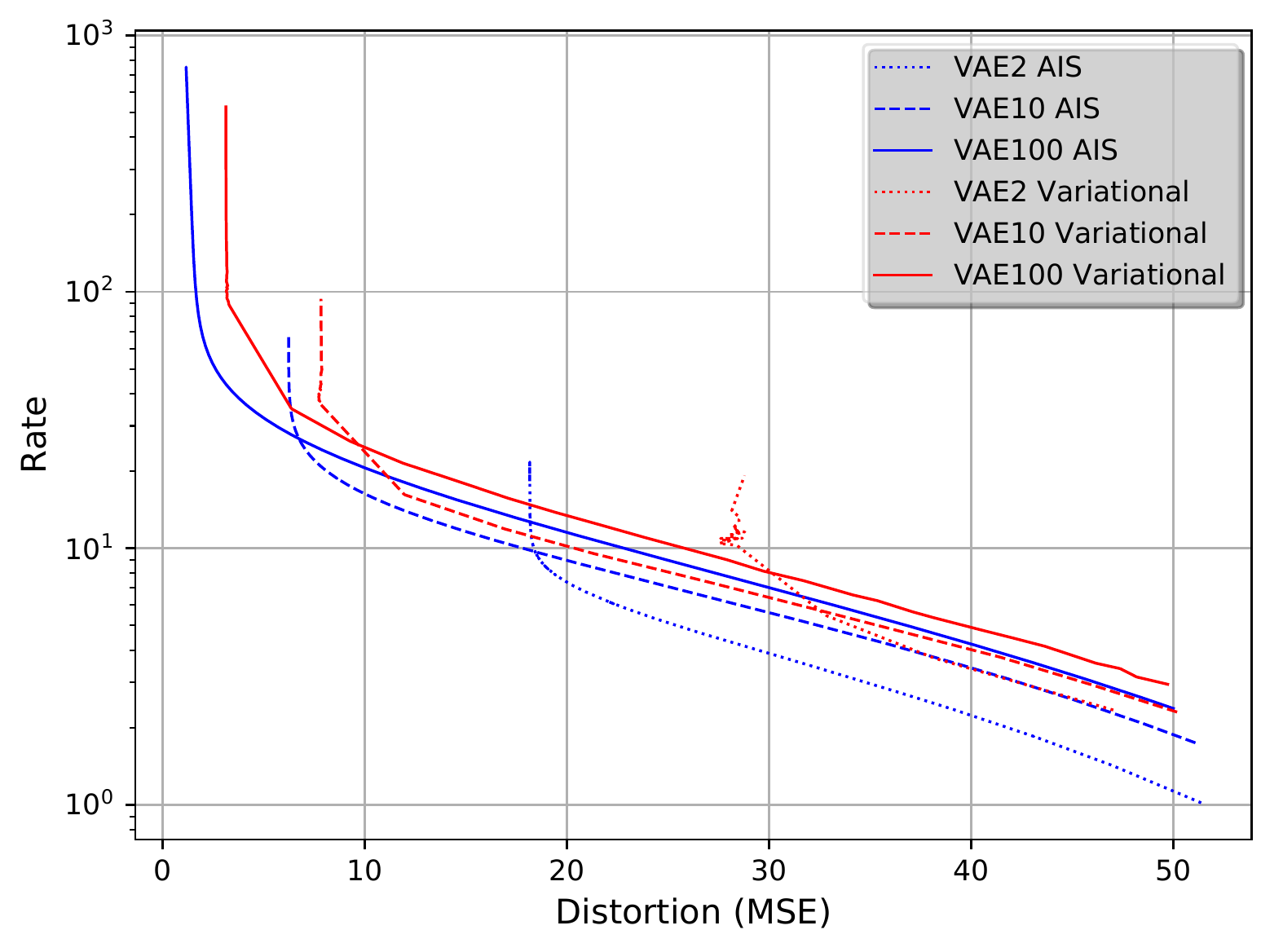}
\vspace{-.3cm}
\caption{\label{fig:var_baseline}Comparison of AIS and variational estimates of variational RD curves. \textbf{(a)} GANs. \textbf{(b)} VAEs.}
\end{figure*}

\vspace{-.3cm}
\subsection{Beyond Pixelwise Mean Squared Error}
\vspace{-.2cm}
\label{sec:deep_feature}
The experiments discussed above all used pixelwise MSE as the distortion metric. However, for natural images, one could use more perceptually valid distortion metrics such as SSIM~\citep{wang2004image}, MSSIM~\citep{wang2003multiscale}, or distances between deep features of a CNN~\citep{johnson2016perceptual}. \myfig{deep_feature_mnist} shows the RD curves of GANs, VAEs, and AAEs on the MNIST, using the MSE on the deep features of a CNN as distortion metric. In all cases, the qualitative behavior of the RD curves with this distortion metric closely matches the qualitative behaviors for pixelwise MSE. We can see from \myfiga{deep_feature_mnist}{a} that similar to the RD curves with MSE distortion, GANs with different depths and code sizes have the same low-rate performance, but as the model gets deeper and wider, the RD curves are pushed down and extended to the left. Similarly, we can see from \myfiga{deep_feature_mnist}{b} that compared to GANs and AAEs, VAEs generally have a better high-rate performance, but worse low-rate performance. The fact that the qualitative behaviors of RD curves with this metric closely match those of pixelwise MSE indicates that the results of our analysis are not overly sensitive to the particular choice of distortion metric.

\vspace{-.3cm}
\subsection{Comparison of AIS and Variational RD Curves}
\vspace{-.1cm}
\label{sec:var}
In this section, we compare the AIS estimate with the variational estimate of variational RD curves of GANs and VAEs. We conducted an experiment where we trained an amortized variational inference network to estimate the rate and distortion at different values of $\beta$ (\myfig{var_baseline}).
Since AIS obtains an upper bound on the RD curve (\myprop{4}), we know from this figure that the AIS estimate is more accurate than the variational estimate for both VAEs and GANs.
In the case of GANs (\myfiga{var_baseline}{a}), we can see that the variational method completely fails to provide any useful bounds and even predicts a wrong ordering for the comparison of GANs, e.g., it predicts GAN-100 is strictly worse than the low-dimensional GANs.
In the case of VAEs (\myfiga{var_baseline}{b}), the AIS outperforms the variational method, but by a smaller margin. We believe this is because VAEs learn a model whose true posterior could fit to the factorized Gaussian approximation of the posterior, and thus variational approximation could also provide useful bounds for estimating the RD curves. However, in the case of GANs the true posterior is highly multi-modal and thus variational approximations fail to provide useful bounds. 

\vspace{-.4cm}
\section{Conclusion}
\vspace{-.2cm}
In this work, we studied rate distortion approximations for evaluating different generative models such as VAEs, GANs and AAEs. We showed that rate distortion curves provide more insights about the model than the log-likelihood alone while requiring roughly the same computational cost. For instance, we observed that while VAEs with larger code size can generally achieve better lossless compression rates, their performances drop at lossy compression in the low-rate regime. Conversely, expanding the capacity of GANs appears to bring substantial reductions in distortion at the high-rate regime without any corresponding deterioration in quality in the low-rate regime. This may help explain the success of large GAN architectures~\citep{brock2018large,karras2018progressive,karras2018style}. We also found that increasing the capacity of GANs by increasing the code size (width) has a very different effect than increasing the depth. The former extends the RD curves leftwards, while the latter pushes the curves down.
Overall, lossy compression yields a richer and more complete picture of the distribution modeling performance of generative models. The ability to quantitatively measure performance tradeoffs should lead to algorithmic insights which can improve these models.

\section{Acknowledgements}
Alireza Makhzani and Roger Grosse acknowledge support from the CIFAR Canadian AI Chairs program. Sicong Huang acknowledges the support related to computing from JinSung Kang at Borealis AI. 
\bibliography{icml20}
\bibliographystyle{icml2020}

\clearpage
\onecolumn
\icmltitlerunning{Appendix: Evaluating Lossy Compression Rates of Deep Generative Models}
\icmltitle{Appendix: Evaluating Lossy Compression Rates of Deep Generative Models}

\begin{icmlauthorlist}
\icmlauthor{Sicong Huang}{equal,uoft,vector,borealis}
\icmlauthor{Alireza Makhzani}{equal,uoft,vector}
\icmlauthor{Yanshuai Cao}{borealis}
\icmlauthor{Roger Grosse}{uoft,vector}
\end{icmlauthorlist}

\printAffiliationsAndNoticeAppendix{\icmlEqualContribution}

\begin{appendices}

\section{Proofs}
\label{app:proof}

\subsection{Proof of \myprop{1}.}
\label{app:proof_1}
\parhead{Proof of \mypropa{1}{a}.}
As $D$ increases, $\mathcal{R}_p(D)$ is minimized over a larger set, so $\mathcal{R}_p(D)$ is non-increasing function of~$D$. 

The distortion $\mathbb{E}_{q(\mathbf{x},\mathbf{z})}[d(\mathbf{x},f(\mathbf{z}))]$ is a linear function of the channel conditional distribution $q(\mathbf{z}|\mathbf{x})$.
The mutual information is a convex function of $q(\mathbf{z}|\mathbf{x})$. The $\mathrm{KL}( q(\mathbf{z})\|p(\mathbf{z}))$ is also convex function of $q(\mathbf{z})$, which itself is a linear function of $q(\mathbf{z}|\mathbf{x})$. Thus $\mathrm{KL}(q(\mathbf{z}|\mathbf{x})\|p(\mathbf{z}))$ is a convex function of $q(\mathbf{z}|\mathbf{x})$.
Suppose for the distortions $D_1$ and $D_2$, $q_1(\mathbf{z}|\mathbf{x})$ and $q_2(\mathbf{z}|\mathbf{x})$ achieve the optimal rates in \myeq{rd} respectively. Suppose the conditional $q_\lambda(\mathbf{z}|\mathbf{x})$ is defined as $q_\lambda(\mathbf{z}|\mathbf{x})=\lambda q_1(\mathbf{z}|\mathbf{x})+(1-\lambda) q_2(\mathbf{z}|\mathbf{x})$.  The $\mathrm{KL}(q(\mathbf{z}|\mathbf{x})\|p(\mathbf{z}))$ objective that the conditional $q_\lambda(\mathbf{z}|\mathbf{x})$ achieves is $\mathcal{I}_\lambda(\mathbf{z};\mathbf{x}) +\mathrm{KL}( q_\lambda(\mathbf{z})\|p(\mathbf{z}))$, and the distortion $D_\lambda$ that this conditional achieves is $D_\lambda = \lambda D_1 +(1-\lambda) D_2$. Now we have
\begin{align}
\mathcal{R}_p(D_\lambda) &\leq \mathcal{I}_\lambda(\mathbf{z};\mathbf{x}) +\mathrm{KL}( q_\lambda(\mathbf{z})\|p(\mathbf{z}))\\
&\leq \lambda \mathcal{I}_1(\mathbf{z};\mathbf{x}) + \lambda \mathrm{KL}( q_1(\mathbf{z})\|p(\mathbf{z})) + (1-\lambda) \mathcal{I}_2(\mathbf{z};\mathbf{x}) + (1-\lambda) \mathrm{KL}( q_2(\mathbf{z})\|p(\mathbf{z})) \\
&= \lambda \mathcal{R}_p(D_1) + (1-\lambda) \mathcal{R}_p(D_2)
\end{align}
which proves the convexity of $\mathcal{R}_p(D)$.

\parhead{Alternative Proof of \mypropa{1}{a}.} We know that $\mathbb{E}_{p_d(\mathbf{x})}\mathrm{KL}( q(\mathbf{z|x})\|p(\mathbf{z}))$ is a convex function of $q(\mathbf{z}|\mathbf{x})$, and $\mathbb{E}_{q(\mathbf{x},\mathbf{z})}[d(\mathbf{x},f(\mathbf{z}))]$ is a linear and thus convex function of $q(\mathbf{z}|\mathbf{x})$. As the result, the following optimization problem is a convex optimization problem.

\vspace{-.8cm}
\begin{align}\label{eq:rpd_perturb}
\min\limits_{q(\mathbf{z}|\mathbf{x})} &\mathbb{E}_{p_d(\mathbf{x})}\mathrm{KL}( q(\mathbf{z|x})\|p(\mathbf{z})) \quad
s.t.\ \mathbb{E}_{q(\mathbf{x},\mathbf{z})}[d(\mathbf{x},f(\mathbf{z}))]\leq 0. 
\end{align}
\vspace{-.5cm}

The rate distortion function $\mathcal{R}_p(D)$ is the perturbation function of the convex optimization problem of \myeq{rpd_perturb}. The convexity of $\mathcal{R}_p(D)$ follows from the fact that the perturbation function of any convex optimization problem is a convex function~\citep{boyd2004convex}.

\parhead{Proof of \mypropa{1}{b}.}
We have
\begin{align}
\min_{p(\mathbf{z})} \mathcal{R}_p(D)
&=
\min_{p(\mathbf{z})}
\,\min_{q(\mathbf{z}|\mathbf{x}):\mathbb{E}[d(\mathbf{x},f(\mathbf{z}))] \leq D}
\,\mathcal{I}(\mathbf{x};\mathbf{z}) +\mathrm{KL}( q(\mathbf{z})\|p(\mathbf{z}))\\
&=
\min_{q(\mathbf{z}|\mathbf{x}):\mathbb{E}[d(\mathbf{x},f(\mathbf{z}))] \leq D}
\,\min_{p(\mathbf{z})}
\,\mathcal{I}(\mathbf{x};\mathbf{z}) +\mathrm{KL}( q(\mathbf{z})\|p(\mathbf{z}))\label{eq:min_min}\\
&=
\min_{q(\mathbf{z}|\mathbf{x}):\mathbb{E}[d(\mathbf{x},f(\mathbf{z}))] \leq D}
\,\mathcal{I}(\mathbf{x};\mathbf{z})\label{eq:min_kl}\\
&=
\mathcal{R}(D).
\end{align}
where in \myeq{min_min}, we have used the fact that for any function $f(x,y)$, we have
\begin{align}
\min_{x}\min_{y}f(x,y)=\min_{y}\min_{x}f(x,y)=\min_{x,y}f(x,y),
\end{align}
and in \myeq{min_kl}, we have used the fact that $\mathrm{KL}( q(\mathbf{z})\|p(\mathbf{z}))$ is minimized when $p(\mathbf{z})=q(\mathbf{z})$.

\parhead{Proof of \mypropa{1}{c}.}
In \mypropa{1}{a}, we showed that $\mathrm{KL}(q(\mathbf{z}|\mathbf{x})\|p(\mathbf{z}))$ is a convex function of $q(\mathbf{z}|\mathbf{x})$, and that the distortion is a linear function of $q(\mathbf{z}|\mathbf{x})$. So the summation of them in \myeq{rdl} will be a convex function of $q(\mathbf{z}|\mathbf{x})$. The unique global optimum of this convex optimization can be found by rewriting \myeq{rdl} as
\begin{align}\label{eq:rpd-proof}
&\mathrm{KL}(q(\mathbf{z|x})\|p(\mathbf{z})) +\beta \mathbb{E}_{q(\mathbf{z}|\mathbf{x})}[d(\mathbf{x},f(\mathbf{z}))] = \mathrm{KL} \big(q(\mathbf{z}|\mathbf{x})\|{1\over Z(\mathbf{x})}p(\mathbf{z})\exp(-\beta d(\mathbf{x},f(\mathbf{z})))\big)-\log Z_\beta(\mathbf{x})
\end{align}
where $Z_\beta(\mathbf{x}) = \int p(\mathbf{z})\exp(-\beta d(\mathbf{x},f(\mathbf{z}))) d\mathbf{z}$.
The minimum of \myeq{rpd-proof} is obtained when the KL divergence is zero. Thus the optimal channel conditional is
\begin{align}
q_\beta^{*}(\mathbf{z}|\mathbf{x})={1\over Z_\beta(\mathbf{x})}p(\mathbf{z})\exp(-\beta d(\mathbf{x},f(\mathbf{z}))).
\end{align}

\subsection{Proof of \myprop{2}.} 
\label{app:proof_2}
\parhead{Proof of \mypropa{2}{a}.}
$\mathcal{R}(D) \leq \mathcal{R}_p(D)$ was proved in \mypropa{1}{b}. To prove the first inequality, note that the summation of rate and distortion is
\begin{align}\label{eq:mi-rewrite}
\mathcal{R}_p(D) + D &= \mathcal{I}(\mathbf{z};\mathbf{x}) + \mathbb{E}_{q^*(\mathbf{x},\mathbf{z})}[-\log p(\mathbf{x|\mathbf{z}})] \\
&=\mathcal{H}_d + \mathbb{E}_{q^*(\mathbf{z})}\mathrm{KL}(q^*(\mathbf{x}|\mathbf{z})\|p(\mathbf{x}|\mathbf{z})) \geq \mathcal{H}_d.
\end{align}
where $q^*(\mathbf{x,z})$ is the optimal joint channel conditional, and $q^*(\mathbf{z})$ and $q^*(\mathbf{x}|\mathbf{z})$ are its marginal and conditional.
The equality happens if there is a joint distribution $q(\mathbf{x},\mathbf{z})$, whose conditional $q(\mathbf{x}|\mathbf{z})=p(\mathbf{x}|\mathbf{z})$, and whose marginal over $\mathbf{x}$ is $p_d(\mathbf{x})$. But note that such a joint distribution might not exist for an arbitrary $p(\mathbf{x}|\mathbf{z})$. 

\parhead{Proof of \mypropa{2}{b}.}
The proof can be easily obtained by using $d(\mathbf{x},f(\mathbf{z}))=-\log p(\mathbf{x}|\mathbf{z})$ in \mypropa{1}{c}.

\parhead{Proof of \mypropa{2}{c}.}
Based on \mypropa{2}{b}, at $\beta=1$, we have
\begin{align}
Z^{*}_{\beta}(\mathbf{x})=\int p(\mathbf{z})p(\mathbf{x}|\mathbf{z}) d\mathbf{z}=p(\mathbf{x}).
\end{align}

\subsection{Proof of \myprop{3}.}
\label{app:proof_3}
The set of pairs of $\big(R_{k}^{\text{AIS}}(\mathbf{x}),D_{k}^{\text{AIS}}(\mathbf{x})\big)$ are achievable variational rate distortion pairs (achieved by $q_{k}^{\text{AIS}}(\mathbf{z}|\mathbf{x})$). Thus, by the definition of $\mathcal{R}_{p}(D)$, $\mathcal{R}^{\text{AIS}}_{p}(D)$ falls in the achievable region of $\mathcal{R}_{p}(D)$ and, thus maintains an upper bound on it: $\mathcal{R}^{\text{AIS}}_{p}(D) \geq \mathcal{R}_{p}(D)$.

\subsection{Proof of \myprop{4}.}
\label{app:proof_4}
AIS has the property that for any step $k$ of the algorithm, the set of chains up to step $k$, and the partial computation of their weights, can be viewed as the result of a complete run of AIS with target distribution $q_{k}^*(\mathbf{z}|\mathbf{x})$. Hence, we assume without loss of generality that we are looking at a complete run of AIS (but our analysis applies to the intermediate distributions as well). 

Let $q_{k}^{\text{AIS}}(\mathbf{z}|\mathbf{x})$ denote the distribution of final samples produced by AIS. More precisely, it is a distribution encoded by the following procedure:
\begin{enumerate}

\item For each data point $\mathbf{x}$, we run $M$ independent AIS chains, numbered $i=1, \ldots, M$. Let $\mathbf{z'}^i_k$ denotes the $k$-th state of the $i$-th chain.
The joint distribution of the forward pass up to the $k$-th state is denoted by $q_f(\mathbf{z'}^i_1, \ldots, \mathbf{z'}^i_k|\mathbf{x})$.
The un-normalized joint distribution of the backward pass is denoted by
\begin{align}\nonumber
\tilde{q}_b(\mathbf{z'}^i_1, \ldots, \mathbf{z'}^i_k|\mathbf{x})= p(\mathbf{z'}^i_k)\exp(-\beta_k d(\mathbf{x}, f(\mathbf{z'}^i_k))) {q_b}(\mathbf{z'}^i_1, \ldots, \mathbf{z'}^i_{k-1}|\mathbf{z'}^i_k, \mathbf{x}).
\end{align}

\item Compute the importance weights and normalized importance weights of each chain using 
\begin{align}
w^{i}_k = \frac{\tilde{q}_b(\mathbf{z'}^i_1, \ldots, \mathbf{z'}^i_k|\mathbf{x})}{q_f(\mathbf{z'}^i_1, \ldots, \mathbf{z'}^i_k|\mathbf{x})} \quad \text{and} \quad \tilde{w}^{i}_k = \frac{{w}^{i}_k}{\sum_{i=1}^{M} {w}^{i}_k}.
\end{align}

\item Select a chain index $S$ with probability of $\tilde{w}^{i}_k$.
\item Assign the selected chain values to $(\mathbf{z}^1_1, \ldots, \mathbf{z}^1_{k})$:
\begin{footnotesize}
\begin{align}
\label{eq:z_selected}
(\mathbf{z}^1_1, \ldots, \mathbf{z}^1_{k}) = (\mathbf{z'}^S_1, \ldots, \mathbf{z'}^S_{k}).
\end{align}
\end{footnotesize}
\item Keep the unselected chain values and re-label them as $(\mathbf{z}^{2:M}_1, \ldots, \mathbf{z}^{2:M}_k)$:
\begin{align}
\label{eq:z_unselected}
(\mathbf{z}^{2:M}_1, \ldots, \mathbf{z}^{2:M}_k) = (\mathbf{z'}^{-S}_1, \ldots, \mathbf{z'}^{-S}_k).
\end{align}
where $-S$ denotes the set of all indices except the selected index $S$.

\item Return $\mathbf{z}=\mathbf{z}^1_{k}$.
\end{enumerate}

More formally, the AIS distribution is
\begin{align}
q_{k}^{\text{AIS}}(\mathbf{z}|\mathbf{x}) = \mathbb{E}_{\prod_{i=1}^M q_f(\mathbf{z'}^i_1, \ldots, \mathbf{z'}^i_k|\mathbf{x})}[\sum_{i=1}^M \tilde{w}^{i}_k\delta(\mathbf{z}-\mathbf{z'}^i_k)].
\end{align}

Using the AIS distribution $q_{k}^{\text{AIS}}(\mathbf{z}|\mathbf{x})$ defined as above, we define the AIS distortion $D_{k}^{\text{AIS}}(\mathbf{x})$ and the AIS rate $R_{k}^{\text{AIS}}(\mathbf{x}) = \mathrm{KL}({q_{k}^{\text{AIS}}(\mathbf{z}|\mathbf{x})}\|p(\mathbf{z}))$ as follows:
\begin{align}
D_{k}^{\text{AIS}}(\mathbf{x}) &= \mathbb{E}_{q_{k}^{\text{AIS}}(\mathbf{z}|\mathbf{x})}[d(\mathbf{x}, f(\mathbf{z}))] \\
R_{k}^{\text{AIS}}(\mathbf{x}) &= \mathrm{KL}({q_{k}^{\text{AIS}}(\mathbf{z}|\mathbf{x})}\|p(\mathbf{z})).
\end{align}
In order to estimate $R_{k}^{\text{AIS}}(\mathbf{x})$ and $D_{k}^{\text{AIS}}(\mathbf{x})$, we define 
\begin{align}
\hat{D}^{\text{AIS}}_k(\mathbf{x}) &= \sum_{i=1}^M \tilde{w}_k^i d(\mathbf{x}, f(\mathbf{z'}^i_k)),\\
\hat{Z}^{\text{AIS}}_k(\mathbf{x}) &= {1 \over M} \sum_{i=1}^M w_k^i \label{eq:z_hat}, \\
\hat{R}^{\text{AIS}}_k(\mathbf{x}) &= -\log \hat{Z}^{\text{AIS}}_k(\mathbf{x}) - \beta_k \hat{D}^{\text{AIS}}_k(\mathbf{x}).\label{eq:r_hat}
\end{align}
We would like to prove that
\begin{align}
\mathbb{E}_{\prod_{i=1}^M q_f(\mathbf{z'}^i_1, \ldots, \mathbf{z'}^i_k|\mathbf{x})}[\hat{D}^{\text{AIS}}_k(\mathbf{x})] &= D_{k}^{\text{AIS}}(\mathbf{x}), \label{eq:prop_d}\\
\mathbb{E}_{\prod_{i=1}^M q_f(\mathbf{z'}^i_1, \ldots, \mathbf{z'}^i_k|\mathbf{x})}[\hat{R}^{\text{AIS}}_k(\mathbf{x})] &\geq R_{k}^{\text{AIS}}(\mathbf{x}).\label{eq:prop_r}
\end{align}

The proof of \myeq{prop_d} is straightforward:
\begin{align}
\label{eq:d_unbiased}
D_{k}^{\text{AIS}}(\mathbf{x})
&=\mathbb{E}_{q_{k}^{\text{AIS}}(\mathbf{z}|\mathbf{x})}[d(\mathbf{x}, f(\mathbf{z}))], \\
&=\int{q_{k}^{\text{AIS}}(\mathbf{z}|\mathbf{x})} d(\mathbf{x}, f(\mathbf{z})) d\mathbf{z}, \\
&=\int\mathbb{E}_{\prod_{i=1}^M q_f(\mathbf{z'}^i_1, \ldots, \mathbf{z'}^i_k|\mathbf{x})}[\sum_{i=1}^M \tilde{w}^{i}_k\delta(\mathbf{z}-\mathbf{z'}^i_k)] d(\mathbf{x}, f(\mathbf{z})) d\mathbf{z}, \\
&=\mathbb{E}_{\prod_{i=1}^M q_f(\mathbf{z'}^i_1, \ldots, \mathbf{z'}^i_k|\mathbf{x})} \sum_{i=1}^M \tilde{w}^{i}_k \big[\int \delta(\mathbf{z}-\mathbf{z'}^i_k) d(\mathbf{x}, f(\mathbf{z})) d\mathbf{z}\big], \\
&=\mathbb{E}_{\prod_{i=1}^M q_f(\mathbf{z'}^i_1, \ldots, \mathbf{z'}^i_k|\mathbf{x})} \sum_{i=1}^M \tilde{w}^{i}_k d(\mathbf{x}, f(\mathbf{z'}^i_k)), \\
&=\mathbb{E}_{\prod_{i=1}^M q_f(\mathbf{z'}^i_1, \ldots, \mathbf{z'}^i_k|\mathbf{x})}[\hat{D}^{\text{AIS}}_k(\mathbf{x})].
\end{align}

\myeq{d_unbiased} shows that $\hat{D}^{\text{AIS}}_k(\mathbf{x})$ is an unbiased estimate of $D_{k}^{\text{AIS}}(\mathbf{x})$. We also know $\log \hat{Z}^{\text{AIS}}_k(\mathbf{x})$ obtained by \myeq{z_hat} is the estimate of the log partition function, and by the Jenson's inequality lower bounds in expectation the true log partition function: $\mathbb{E}[\log \hat{Z}_{k}^{\text{AIS}}(\mathbf{x})]\leq \log Z_{k}(\mathbf{x})$. After obtaining $\hat{D}^{\text{AIS}}_k(\mathbf{x})$ and $\log \hat{Z}_{k}^{\text{AIS}}(\mathbf{x})$, we use \myeq{r_hat} to obtain $\hat{R}^{\text{AIS}}_k(\mathbf{x})$. Now, it remains to prove \myeq{prop_r}, which states that $\hat{R}^{\text{AIS}}_k(\mathbf{x})$ upper bounds the AIS rate term $R_{k}^{\text{AIS}}(\mathbf{x})$ in expectation.

Let $q_{k}^{\text{AIS}}(\mathbf{z}^{1:M}_1, \ldots, \mathbf{z}^{1:M}_k|\mathbf{x})$ denote the joint AIS distribution over all states of $\{\mathbf{z}^{1:M}_1, \ldots, \mathbf{z}^{1:M}_k\}$, defined in \myeq{z_selected} and \myeq{z_unselected}.
It can be shown that (see~\citet{domke2018importance})
\begin{align}
q_{k}^{\text{AIS}}(\mathbf{z}^{1:M}_1, \ldots, \mathbf{z}^{1:M}_k | \mathbf{x})
&= \frac{\tilde{q}_b(\mathbf{z}^1_1, \ldots, \mathbf{z}^1_{k}|\mathbf{x}) \, \prod_{i=2}^M q_f(\mathbf{z}^i_1, \ldots, \mathbf{z}^i_{k}|\mathbf{x})}{\hat{Z}^{\text{AIS}}_k(\mathbf{x})} \\
&= \frac{p(\mathbf{z}^1_k)\exp(-\beta_k d(\mathbf{x}, f(\mathbf{z}^1_k))) \, {q_b}(\mathbf{z}^1
_1, \ldots, \mathbf{z}^1_{k-1}|\mathbf{z}^1_k, \mathbf{x}) \, \prod_{i=2}^M  q_f(\mathbf{z}^i_1, \ldots, \mathbf{z}^i_{k}|\mathbf{x})}     {\hat{Z}^{\text{AIS}}_k(\mathbf{x})}\label{eq:ais_joint}
\end{align}
In order to simplify notation, suppose $\mathbf{z}^1_k$ is denoted by $\mathbf{z}$, and all the other variables $\{\mathbf{z}^{1:M}_{1}, \ldots, \mathbf{z}^{1:M}_{k-1}, \mathbf{z}^{2:M}_k\}$ are denoted by $\mathbf{V}$. Using this notation, we define $p(\mathbf{V}|\mathbf{z},\mathbf{x})$ and $q_{k}^{\text{AIS}}(\mathbf{z}, \mathbf{V}|\mathbf{x})$ as follows:
\begin{align}
p(\mathbf{V}|\mathbf{z},\mathbf{x})
&\defeq
{q_b}(\mathbf{z}^1_1, \ldots, \mathbf{z}^1_{k-1}|\mathbf{z}^1_k, \mathbf{x}) \, \prod_{i=2}^M  q_f(\mathbf{z}^i_1, \ldots, \mathbf{z}^i_{k}|\mathbf{x}), \\
q_{k}^{\text{AIS}}(\mathbf{z}, \mathbf{V}|\mathbf{x})
&\defeq
q_{k}^{\text{AIS}}(\mathbf{z}^{1:M}_1, \ldots, \mathbf{z}^{1:M}_k|\mathbf{x}) 
\end{align}
Using the above notation, \myeq{ais_joint} can be re-written as
\begin{align}
\hat{Z}^{\text{AIS}}_k(\mathbf{x}) = \frac{p(\mathbf{z}) \, \exp(-\beta_k d(\mathbf{x}, f(\mathbf{z})))\, p(\mathbf{V} | \mathbf{z}, \mathbf{x})}{q_{k}^{\text{AIS}}(\mathbf{z}, \mathbf{V}|\mathbf{x})}.
\end{align}
Hence,
\begin{equation}
\begin{aligned}
\mathbb{E}[\log \hat{Z}^{\text{AIS}}_k(\mathbf{x})] &= \mathbb{E}[\log p(\mathbf{z}) - \log q_{k}^{\text{AIS}}(\mathbf{z}, \mathbf{V}|\mathbf{x}) + \log p(\mathbf{V}|\mathbf{x}, \mathbf{z})] - \beta_k \mathbb{E}[d(\mathbf{x}, f(\mathbf{z}))] \\
&= -\mathrm{KL}(q_{k}^{\text{AIS}}(\mathbf{z}, \mathbf{V}|\mathbf{x}) \| p(\mathbf{z}) p(\mathbf{V} | \mathbf{z}, \mathbf{x})) - \beta_k \mathbb{E}[d(\mathbf{x}, f(\mathbf{z}))] \\
&\leq -\mathrm{KL}(q_{k}^{\text{AIS}}(\mathbf{z}|\mathbf{x}) \| p(\mathbf{z})) - \beta_k \mathbb{E}[d(\mathbf{x}, f(\mathbf{z}))],
\end{aligned}
\end{equation}
where the inequality follows from the monotonicity of KL divergence. Rearranging terms, we bound the rate:
\begin{align}\label{eq:rate_ub}
R_{k}^{\text{AIS}}(\mathbf{x}) &= \mathrm{KL}(q_{k}^{\text{AIS}}(\mathbf{z}|\mathbf{x}) \| p(\mathbf{z}))
\leq -\mathbb{E}[\log \hat{Z}^{\text{AIS}}_k(\mathbf{x})] - \beta_k \mathbb{E}[d(\mathbf{x}, f(\mathbf{z}))]
= \mathbb{E}[\hat{R}_{k}^{\text{AIS}}(\mathbf{x})].
\end{align}
\myeq{rate_ub} shows that $\hat{R}_{k}^{\text{AIS}}(\mathbf{x})$ upper bounds the AIS rate $R_{k}^{\text{AIS}}(\mathbf{x})$ in expectation. We also showed $\hat{D}^{\text{AIS}}_k(\mathbf{x})$ is an unbiased estimate of the AIS distortion $D_{k}^{\text{AIS}}(\mathbf{x})$. Hence, the estimated AIS rate distortion curve upper bounds the AIS rate distortion curve in expectation: $\mathbb{E}[\hat{\mathcal{R}}^{\text{AIS}}_{p}(D)] \geq \mathcal{R}^{\text{AIS}}_{p}(D)$.

\section{Validation of AIS experiments}
\label{app:experiment_ais}

\subsection{Analytical Solution of the Variational Rate Distortion Optimization on the Linear VAE}
\label{app:analytical}
We compared our AIS results with the analytical solution of the variational rate distortion optimization on a linear VAE trained on MNIST as shown in \myfig{vae_analytical}.

In order to derive the analytical solution, we first find the optimal distribution $q^{*}_{\beta}(\mathbf{z}|\mathbf{x})$ from \mypropa{2}{b}.
For simplicity, we assume a fixed identity covariance matrix $I$ at the output of the conditional likelihood of the linear VAE decoder. In other words, the decoder of the VAE is simply: $\bx=\bW \bz+\bbb+\epsilon$, where $\bx$ is the observation, $\bb{z}$ is the latent code vector, $\bb{W}$ is the decoder weight matrix and $\bbb$ is the bias. The observation noise of the decoder is $\epsilon \sim \mathcal{N} (\bzero,\,\bI)$. It's easy to show that the conditional likelihood raised to a power $\beta$ is: $p(\bx | \bz)^\beta = \mathcal{N} (\bx | \bW \bz + \bbb, \frac{1}{\beta} \bI)$.
Then, $q^{*}_{\beta}(\mathbf{z}|\mathbf{x})=  \mathcal{N} (\bz | \mu_\beta, \bSigma_\beta)$, where
\begin{align}
\bmu_\beta &=  \Exp{q^{*}_{\beta}(\mathbf{z}|\mathbf{x})}{\bz} =\bWT(\bW \bWT + \beta^{-1} \bI)\inv (\bx - \bbb), \\
\bSigma_\beta &= \Cov{q^{*}_{\beta}(\mathbf{z}|\mathbf{x})}{\bz} =\bI - \bWT(\bW \bWT + \beta^{-1} \bI)\inv \bW.
\label{eq:analytical_mean}
\end{align}
For numerical stability, we can further simplify the above by taking the SVD of $\bW$: Suppose we have $\bW = \bU \bD \bVT$. We can use the Woodbury Matrix Identity to the matrix inversion operation to obtain
\begin{align}
\bmu_\beta &=  \bV\bR_{\beta} \bUT (\bx - \bbb),\\
\bSigma_\beta &=\bV\bS_{\beta} \bVT,
\label{eq:svd_analytical}
\end{align}

where $\bR_{\beta}$ is a diagonal matrix with the $i$-th diagonal entry being $\frac{d_i}{d_i^2+\frac{1}{\beta}}$ and $\bS_{\beta}$ is a diagonal matrix with the $i$-th diagonal entry being $\frac{1}{\beta d_i^2+1}$, where $d_i$ is the $i$-th diagonal entry of $\bD$. The analytical solution for optimal rate is:

\begin{align}
 D_{KL}(q^{*}_{\beta}(\mathbf{z}|\mathbf{x}) || p(\bz))  &=   D_{KL}(\mathcal{N} (\bz | \bmu_\beta, \bSigma_\beta) || \mathcal{N} (\bz | \bzero, \bI))\\
 &= \frac {1}{2}\left(\operatorname {tr} \left( \bSigma_\beta\right)+(-\bmu_\beta)^{\mathsf {T}} (-\bmu_\beta)-k+\ln \left({{(\det \bSigma_\beta)^{-1}}}\right)\right)\\
 &= \frac {1}{2}\left(\operatorname {tr} \left( \bSigma_\beta\right)+(\bmu_\beta)^{\mathsf {T}} (\bmu_\beta)-k-\ln \left({\det \bSigma_\beta}\right)\right)
\label{eq:lienar_rate}\end{align}
Where k is the dimension of the latent code $\bz$. With  negative log-likelihood as the distortion metric, the analytically form of distortion term is:
\begin{align}
\Exp{q^{*}_{\beta}(\mathbf{z}|\mathbf{x})}{-\log p(\bx | \bz)} &= \int_{-\infty}^{\infty} -\log((2 \pi)^{-k/2} \exp \big\{- \frac{1}{2}(\bx - (\bW \bz+\bbb) )\tp (\bx - (\bW \bz+\bbb))\big\}) q^{*}_{\beta}(\mathbf{z}|\mathbf{x}) d\bz  \\
&= -(\log({(2 \pi)^{-k/2}})+ \frac{1}{2}\int_{-\infty}^{\infty} \big\{(\bx - (\bW \bz+\bbb) )\tp (\bx - (\bW \bz+\bbb))\big\} q^{*}_{\beta}(\mathbf{z}|\mathbf{x}) d\bz \\
&=\frac{k}{2}\log(2 \pi) + \frac{1}{2} (\bx - \bbb) \tp (\bx - \bbb ) -  (\bW  \bmu_\beta) \tp (\bx - \bbb) + \frac{1}{2}  \Exp{q^{*}_{\beta}(\mathbf{z}|\mathbf{x})}{(\bW \bz)\tp (\bW \bz)} 
\label{eq:analytical_distortion}
\end{align}
where $\Exp{q^{*}_{\beta}(\mathbf{z}|\mathbf{x})}{(\bW \bz)\tp (\bW \bz)}$ can be obtained by change of variable: Let $\by=\bW \bz$, then: 
\begin{align}
\Exp{q^*(y)}{\by}&=\bW \bmu_\beta = \bU(\bI - \bS_\beta) \bUT (\bx - \bbb) \\
\Cov{q^*(y)}{\by} &= \bW \bSigma_\beta \bWT = \bU \bD \bS_\beta \bD \bUT \\
\Exp{q^{*}_{\beta}(\mathbf{z}|\mathbf{x})}{(\bW \bz)\tp (\bW \bz)} &= \Exp{q^*(y)}{\by \tp \by}
= \Exp{q^*(y)}{\by} \tp \Exp{q^*(y)}{\by} +  \operatorname {tr}(\Cov{q^*(y)}{\by})
\label{eq:Eqwz}\end{align}

\subsection{The BDMC Gap}
\label{app:bdmc}
\begin{figure*}[t]
\centering
\subfigure[VAEs]{
\includegraphics[scale=.34]{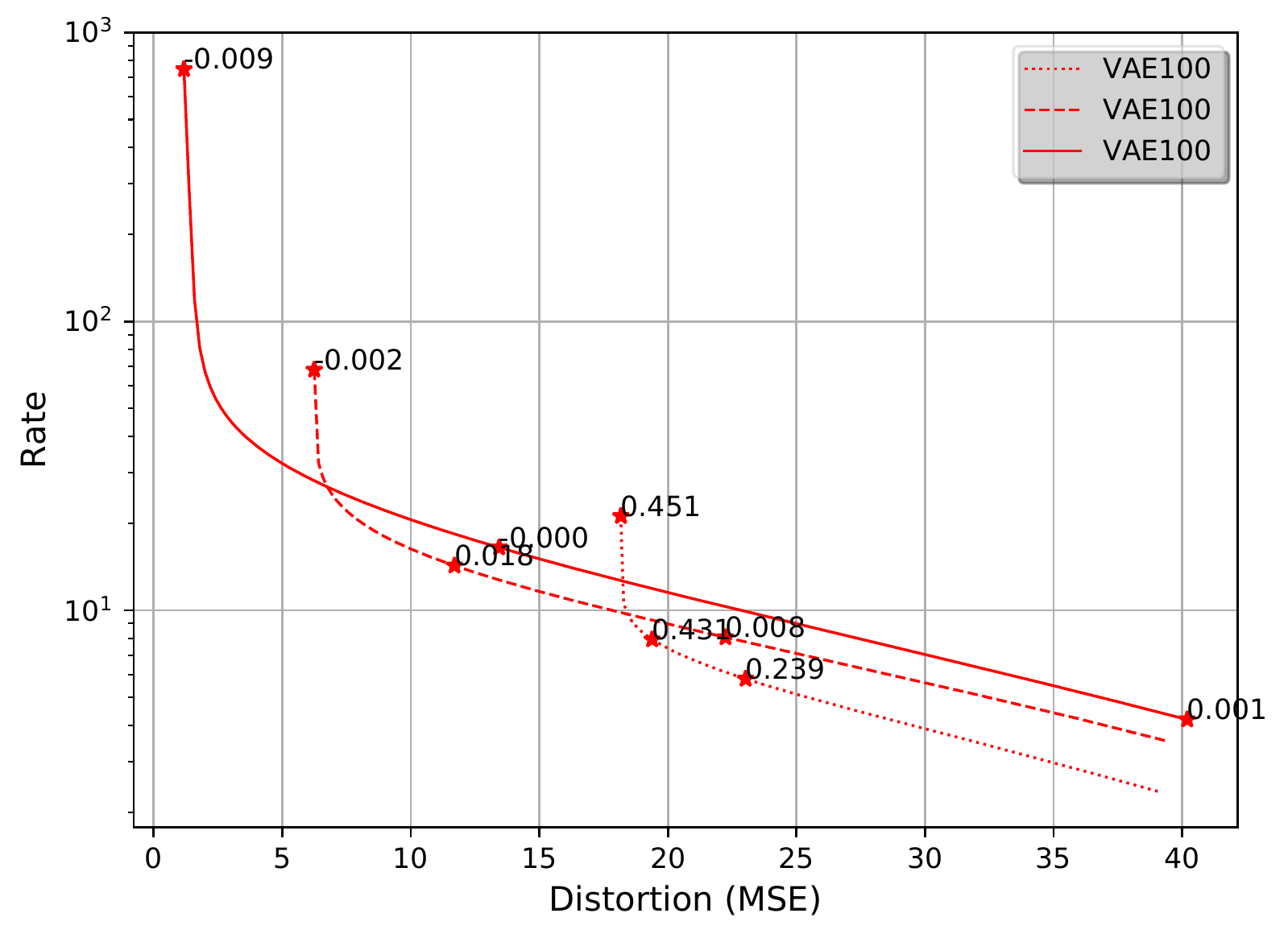}}
\subfigure[GANs]{
\includegraphics[scale=.34]{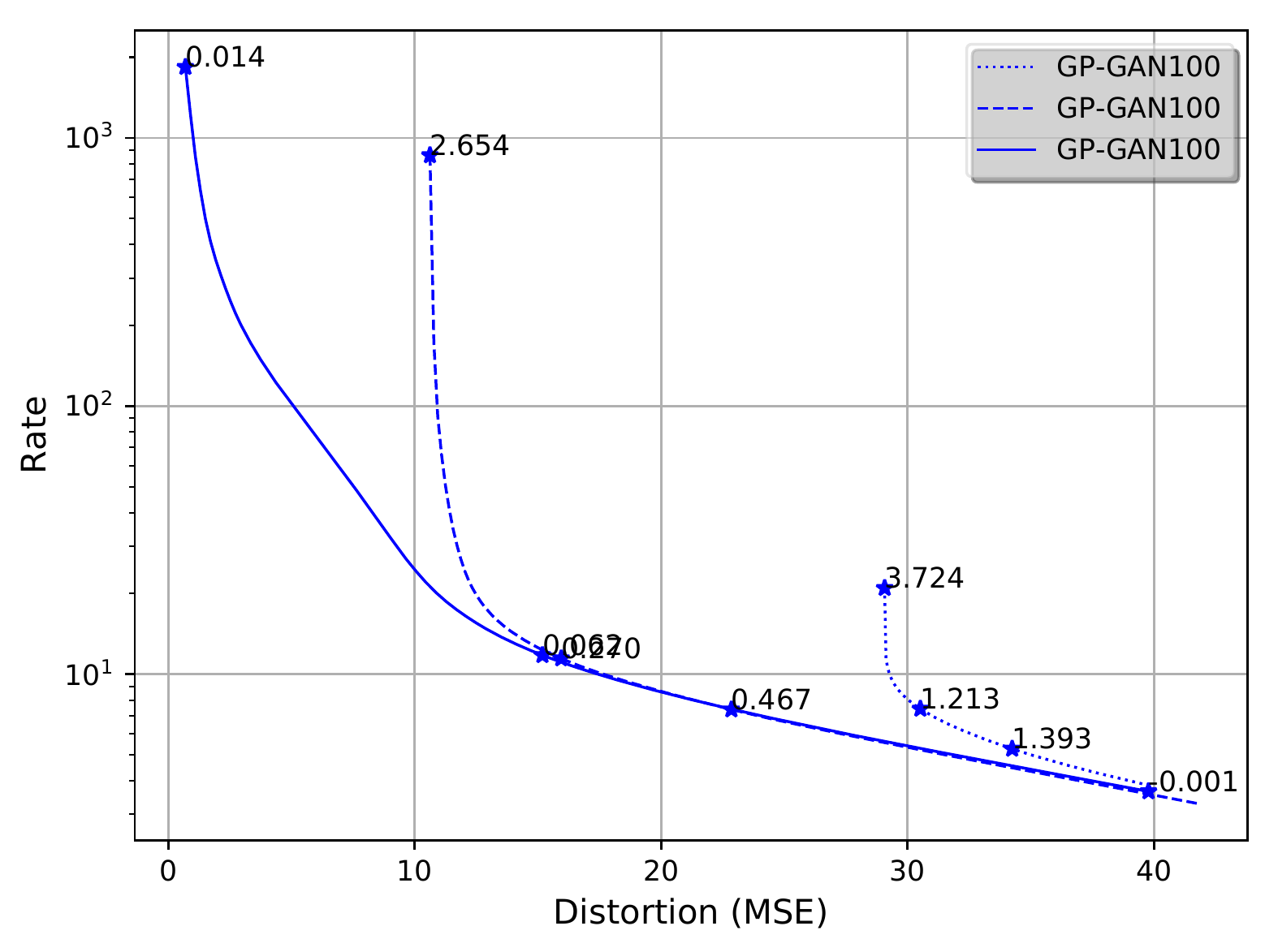}}
\subfigure[AAEs]{
\includegraphics[scale=.34]{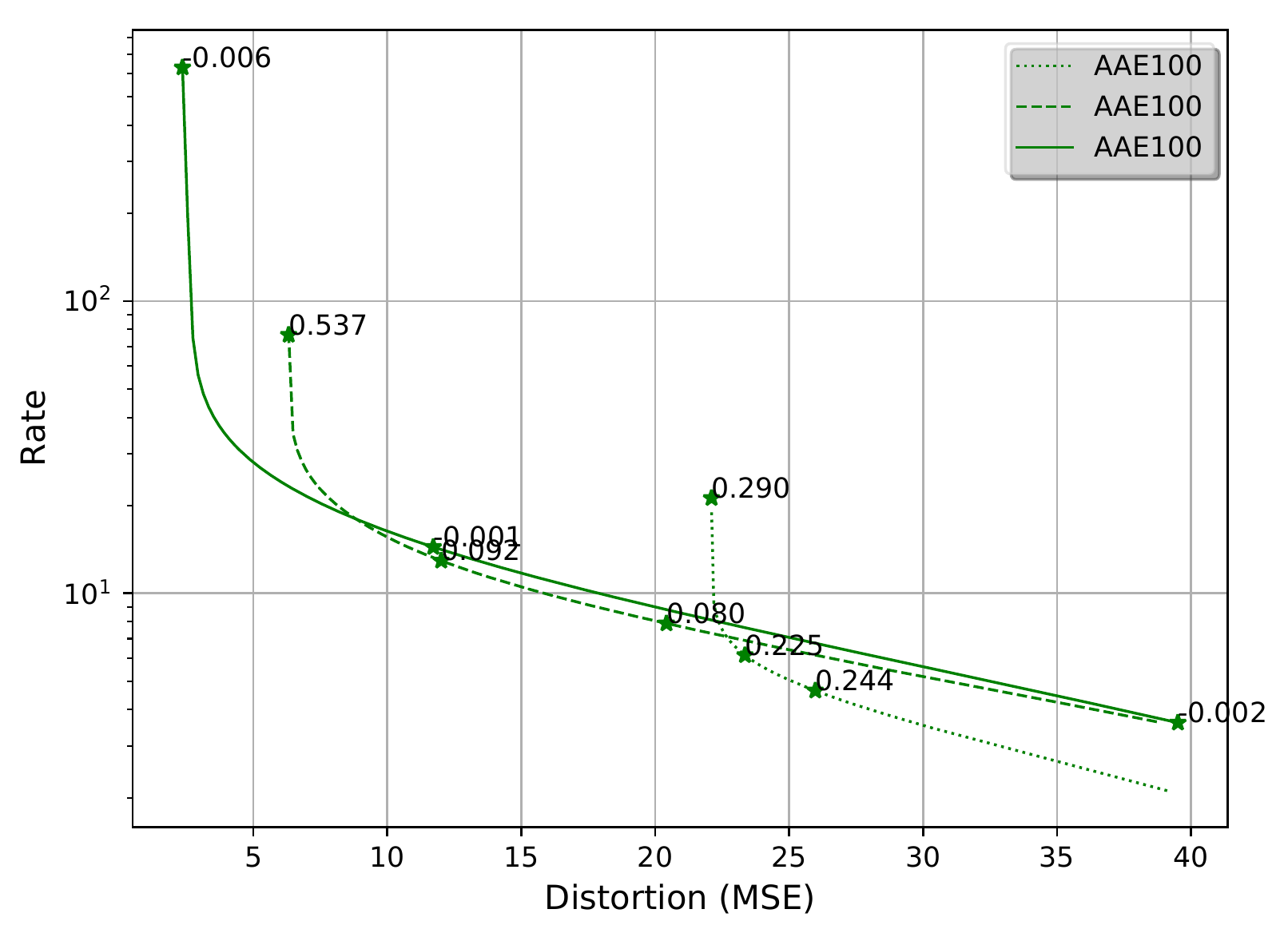}}
\caption{\label{fig:bdmc} The BDMC gaps annotated on estimated AIS Variational Rate Distortion curves of (a) VAEs, (b) GANs, and (c) AAEs.}
\end{figure*}

We evaluated the tightness of the AIS estimate by computing the BDMC gaps using the same AIS settings. \myfig{bdmc}, shows the BDMC gaps at different compression rates for the VAE, GAN and AAE experiments on the MNIST dataset. The largest BDMC gap for VAEs and AAEs was 0.537 nats, and the largest BDMC gap for GANs was 3.724 nats, showing that our AIS upper bounds are tight.

\section{Experimental Details}
The code for reproducing all the experiments of this paper can be found at:
 \href{https://github.com/huangsicong/rate_distortion}{https://github.com/huangsicong/rate\_distortion}. 
\label{app:exp}

\subsection{Datasets and Models}
\label{app:models}
We used MNIST~\citep{lecun1998gradient} and CIFAR-10~\citep{krizhevsky2009learning} datasets in our experiments.

\parhead{Real-Valued MNIST.} For the VAE experiments on the real-valued MNIST dataset (\myfiga{vae_gan_mnist}{a}), we used the ``VAE-50'' architecture described in~\citep{wu2016quantitative}, and only changed the code size in our experiments. The decoder variance is a global parameter learned during the training. The network was trained for 1000 epochs with the learning rate of 0.0001 using the Adam optimizer~\citep{adam}, and the checkpoint with the best validation loss was used for the rate distortion evaluation.  

For the GAN experiments on MNIST (\myfiga{gan_mnist_cifar}{a}), we used the ``GAN-50'' architecture described in~\citep{wu2016quantitative}. In order to stabilize the training dynamic, we used the gradient penalty (GP)~\citep{salimans2016improved}. In our deep architectures, we used code sizes of $d\in\{2,5,10,100\}$ and three hidden layers each having $1024$ hidden units to obtain the following GAN models: Deep-GAN2, Deep-GAN5, Deep-GAN10 and Deep-GAN100. The shallow GANs architectures are similar to the deep architectures but with one layer of hidden units.

\parhead{CIFAR-10.} For the CIFAR-10 experiments (\myfiga{gan_mnist_cifar}{b}), we experimented with different GAN models such as DCGAN~\citep{radford2015unsupervised}, DCGAN with Gradient Penalty (GP-GAN)~\citep{gulrajani2017improved}, Spectral Normalization (SN-GAN)~\citep{miyato2018spectral}, and 
DCGAN with Binarized Representation Entropy Regularization {(BRE-GAN)}~\citep{BRE2018}. The numbers at the end of each GAN name in \myfiga{gan_mnist_cifar}{b} indicate the code size.  

\subsection{AIS Settings for RD Curves}
\label{app:settings}

We evaluated each RD curve at 2000 points corresponding to different values of $\beta$, with $N\gg2000$ intermediate distributions. We used a sigmoid temperature schedule as used in~\citet{wu2016quantitative}. We used $\beta_\text{max} \approx 3000$ for 100 dimensional models (GAN100, VAE100, and AAE100), and used $\beta_\text{max}\approx 36000$ for the rest of the models (2, 5 and 10 dimensional).
For the 2, 5 and 10 dimensional models, we used $N=60000$ intermediate distributions. For 100 dimensional models, we used $N=1600000$ intermediate distributions in order to obtain small BDMC gaps. We used 20 leap frog steps for HMC, 40 independent chains, on a single batch of 50 images. On the MNIST dataset, we also tested with a larger batch size of 500 MNIST images, but did not observe a significant difference in average rates and distortions. On a P100 GPU, for MNIST, it takes 4-7 hours to compute an RD curve with $N=60000$ intermediate distributions and takes around 7 days for $N=160000$ intermediate distributions. For all of the CIFAR experiments, we used the temperature schedule with $N=60000$ intermediate distributions, and each experiment takes about 7 days to complete.

\begin{figure}[t]
  \begin{center}
    \includegraphics[scale=.4]{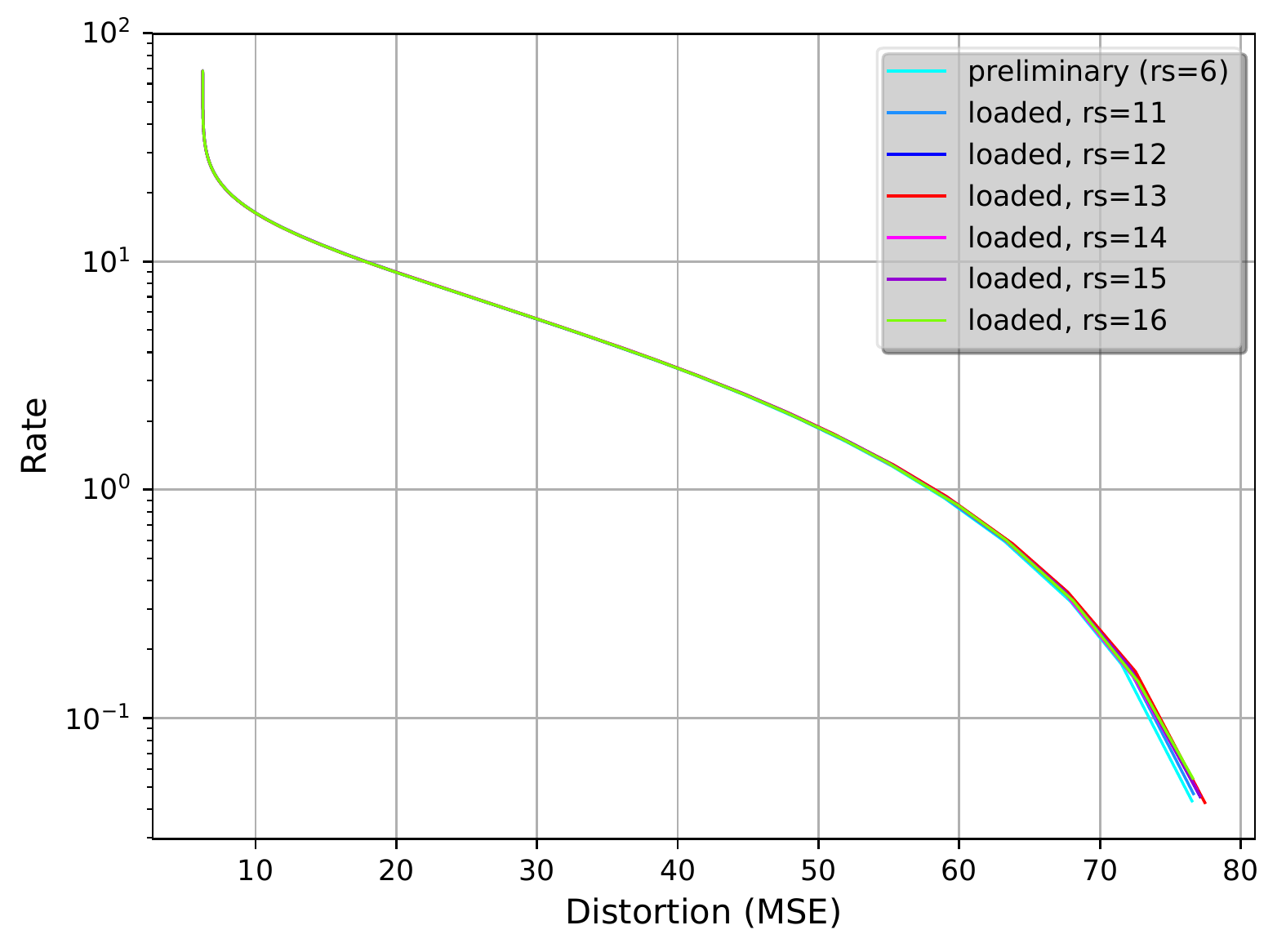}
  \end{center}
  \vspace{-.3cm}
  \caption{\label{fig:load}The variational rate distortion curves obtained by adaptively tuning the HMC parameters in the preliminary run, and pre-loading the HMC parameters in the second formal run. "rs" in the legend indicates the random seed used in the second run.}
\end{figure}

\subsection{Adaptive Tuning of HMC Parameters.}
While running the AIS chain, the parameters of the HMC kernel cannot be adaptively tuned, since it would violate the Markovian property of the chain.
So in order to be able to adaptively tune HMC parameters such as the number of leapfrog steps and the step size, in all our experiments, we first do a preliminary run where the HMC parameters are adaptively tuned to yield an average acceptance probability of $65\%$ as suggested in~\citet{neal2001annealed}.
Then in the second ``formal'' run, we pre-load and fix the HMC parameters found in the preliminary run, and start the chain with a new random seed to obtain our final results. Interestingly, we observed that the difference in the RD curves obtained from the preliminary run and the formal runs with different random seeds is very small, as shown in \myfig{load}. This figure shows that the AIS with the HMC kernel is robust against different choices of random seeds for approximating the RD curve of VAE10.

\section{High-Rate vs. Low-Rate Reconstructions}\label{app:reconst}
\begin{figure}[t]
\centering
\subfigure[Original MNIST test images\hspace{-.2cm}]{
\includegraphics[scale=.6]{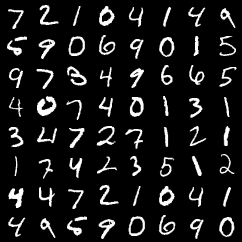}}

\subfigure[Low Rate VAE2]{
\includegraphics[scale=.5]{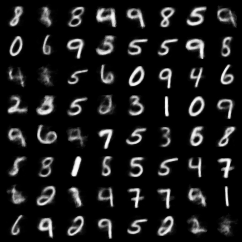}}
\hspace{.25cm}
\subfigure[Low Rate AAE2]{
\includegraphics[scale=.5]{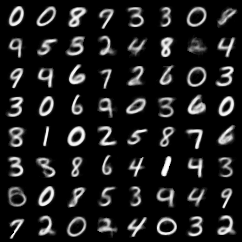}}
\hspace{.25cm}
\subfigure[Low Rate GAN2]{
\includegraphics[scale=.5]{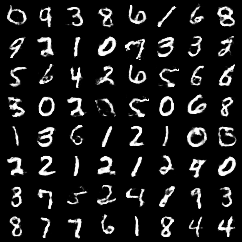}}

\subfigure[Low Rate VAE10]{
\includegraphics[scale=.5]{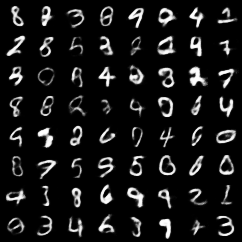}}
\hspace{.25cm}
\subfigure[Low Rate AAE10]{
\includegraphics[scale=.5]{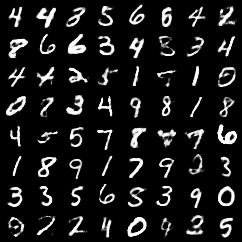}}
\hspace{.25cm}
\subfigure[Low Rate GAN10]{
\includegraphics[scale=.5]{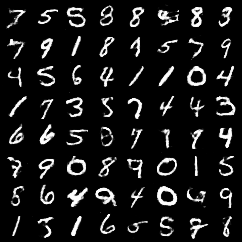}}

\subfigure[Low Rate VAE100]{
\includegraphics[scale=.5]{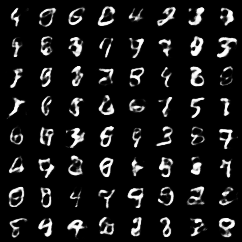}}
\hspace{.25cm}
\subfigure[Low Rate AAE100]{
\includegraphics[scale=.5]{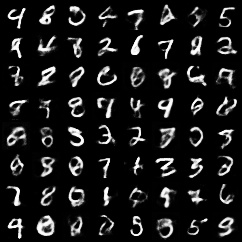}}
\hspace{.25cm}
\subfigure[Low Rate GAN100]{
\includegraphics[scale=.5]{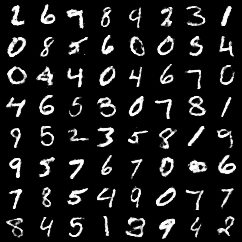}}
\vspace{.25cm}
\caption{\label{fig:low_rate}Low-rate reconstructions ($\beta=0$) of VAEs, GANs and AAEs on MNIST.}
\end{figure}

\begin{figure}[t]
\centering
\subfigure[Original MNIST test images.]{
\includegraphics[scale=.58]{fig/original_mnist}}

\subfigure[High Rate VAE2]{
\includegraphics[scale=.5]{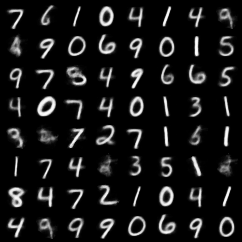}}
\hspace{.25cm}
\subfigure[High Rate AAE2]{
\includegraphics[scale=.5]{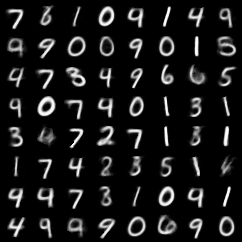}}
\hspace{.25cm}
\subfigure[High Rate GAN2]{
\includegraphics[scale=.5]{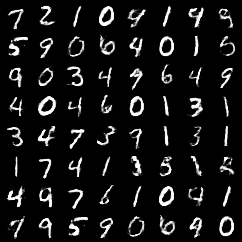}}

\subfigure[High Rate VAE10]{
\includegraphics[scale=.5]{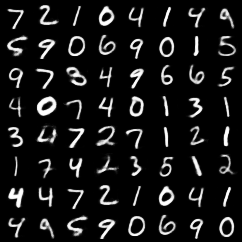}}
\hspace{.25cm}
\subfigure[High Rate AAE10]{
\includegraphics[scale=.5]{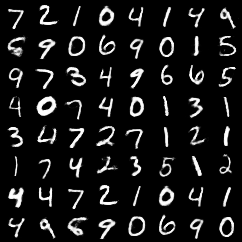}}
\hspace{.25cm}
\subfigure[High Rate GAN10]{
\includegraphics[scale=.5]{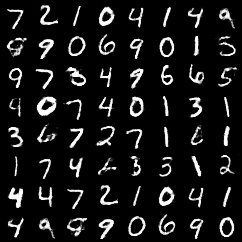}}

\subfigure[High Rate VAE100]{
\includegraphics[scale=.5]{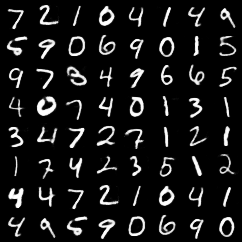}}
\hspace{.25cm}
\subfigure[High Rate AAE100]{
\includegraphics[scale=.5]{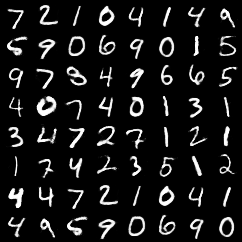}}
\hspace{.25cm}
\subfigure[High Rate GAN100]{
\includegraphics[scale=.5]{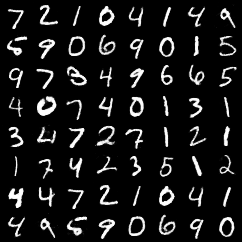}}
\vspace{-.3cm}
\caption{\label{fig:high_rate}High-rate reconstructions ($\beta_\text{max}$) of VAEs, GANs and AAEs on MNIST. $\beta_\text{max}=3333$ for 100 dimensional models, and $\beta_\text{max}=36000$ for the 2 and 10 dimensional models.}
\end{figure}

In this section, we visualize the high-rate ($\beta \approx 3500$) and low-rate ($\beta=0$) reconstructions of the MNIST images for VAEs, GANs and AAEs with different hidden code sizes.
The qualitative results are shown in \myfig{low_rate} and \myfig{high_rate}, which is consistent with the quantitative results presented in the experiment section of the paper.

\end{appendices}
\end{document}

%% file: macros.tex
\usepackage{amsmath}
\usepackage{graphicx}       
\usepackage[titletoc,title]{appendix}
\usepackage{graphics}
\usepackage{subfigure}
\usepackage{float}
\usepackage{hyperref}
\usepackage{xcolor}
\usepackage{wrapfig}
\usepackage[export]{adjustbox}
\usepackage{float}
\usepackage{amssymb}
\usepackage{enumitem}
\usepackage{mathtools}

\hypersetup{
    colorlinks,
    linkcolor={blue!50!black},
    citecolor={blue!50!black},
    urlcolor={blue!50!black}
}

\newcommand{\myprop}[1]{\hyperref[prop:#1]{Prop.~#1}}
\newcommand{\mypropa}[2]{\hyperref[prop:#1]{Prop.~#1#2}}
\newcommand{\mycor}[1]{\hyperref[cor:#1]{Cor. #1}}
\newcommand{\myeq}[1]{\hyperref[eq:#1]{Eq.~\ref*{eq:#1}}}
\newcommand{\mysec}[1]{\hyperref[sec:#1]{Section~\ref*{sec:#1}}}
\newcommand{\mytable}[1]{\hyperref[table:#1]{Table~\ref*{table:#1}}}
\newcommand{\myfig}[1]{\hyperref[fig:#1]{Fig.~\ref*{fig:#1}}}
\newcommand{\myfiga}[2]{\hyperref[fig:#1]{Fig.~\ref*{fig:#1}#2}}
\newcommand{\myfigab}[3]{\hyperref[fig:#1]{Fig.~\ref*{fig:#1}#2,#3}}
\newcommand{\myfigabc}[4]{\hyperref[fig:#1]{Fig.~\ref*{fig:#1}#2,#3,#4}}
\newcommand{\myfigabcd}[5]{\hyperref[fig:#1]{Fig.~\ref*{fig:#1}#2,#3,#4,#5}}
\newcommand{\myapp}[1]{\hyperref[app:#1]{Appendix~\ref*{app:#1}}}

\DeclareRobustCommand{\parhead}[1]{\textbf{#1}~}

\newcommand{\defeq}{\vcentcolon=}

\newcommand{\inv}{^{-1}}
\newcommand{\bb}[1]{\mathbf{#1}}
\newcommand{\bbb}{\bb{b}}
\newcommand{\bx}{\bb{x}}

\newcommand{\by}{\bb{y}}

\newcommand{\bz}{\bb{z}}
\newcommand{\bSigma}{\boldsymbol{\Sigma}}
\newcommand{\bmu}{\boldsymbol{\mu}}

\newcommand{\bzero}{\bb{0}}

\newcommand{\bV}{\bb{V}}
\newcommand{\bW}{\bb{W}}
\newcommand{\bS}{\bb{S}}
\newcommand{\bR}{\bb{R}}
\newcommand{\bU}{\bb{U}}
\newcommand{\bWT}{\bb{W}^\intercal}
\newcommand{\bUT}{\bb{U}^\intercal}
\newcommand{\bVT}{\bb{V}^\intercal}
\newcommand{\bD}{\bb{D}}
\newcommand{\bI}{\bb{I}}

\newcommand{\tp}{^\intercal}

\newcommand{\Exp}[2]{\mathbb{E}_{#1}\left[#2\right]}

\newcommand{\Cov}[2]{\mathrm{Cov}_{#1}\left[#2\right]}

%% file: icml20.bbl
\begin{thebibliography}{45}
\providecommand{\natexlab}[1]{#1}
\providecommand{\url}[1]{\texttt{#1}}
\expandafter\ifx\csname urlstyle\endcsname\relax
  \providecommand{\doi}[1]{doi: #1}\else
  \providecommand{\doi}{doi: \begingroup \urlstyle{rm}\Url}\fi

\bibitem[Alemi et~al.(2018)Alemi, Poole, Fischer, Dillon, Saurous, and
  Murphy]{alemi2018fixing}
Alemi, A., Poole, B., Fischer, I., Dillon, J., Saurous, R.~A., and Murphy, K.
\newblock Fixing a broken elbo.
\newblock In \emph{International Conference on Machine Learning}, pp.\
  159--168, 2018.

\bibitem[Alemi et~al.(2016)Alemi, Fischer, Dillon, and Murphy]{alemi2016deep}
Alemi, A.~A., Fischer, I., Dillon, J.~V., and Murphy, K.
\newblock Deep variational information bottleneck.
\newblock \emph{arXiv preprint arXiv:1612.00410}, 2016.

\bibitem[Arimoto(1972)]{arimoto1972algorithm}
Arimoto, S.
\newblock An algorithm for computing the capacity of arbitrary discrete
  memoryless channels.
\newblock \emph{IEEE Transactions on Information Theory}, 18\penalty0
  (1):\penalty0 14--20, 1972.

\bibitem[Arjovsky et~al.(2017)Arjovsky, Chintala, and
  Bottou]{arjovsky2017wasserstein}
Arjovsky, M., Chintala, S., and Bottou, L.
\newblock Wasserstein {GAN}.
\newblock \emph{arXiv preprint arXiv:1701.07875}, 2017.

\bibitem[Ball{\'e} et~al.(2018)Ball{\'e}, Minnen, Singh, Hwang, and
  Johnston]{balle2018variational}
Ball{\'e}, J., Minnen, D., Singh, S., Hwang, S.~J., and Johnston, N.
\newblock Variational image compression with a scale hyperprior.
\newblock \emph{arXiv preprint arXiv:1802.01436}, 2018.

\bibitem[Boyd \& Vandenberghe(2004)Boyd and Vandenberghe]{boyd2004convex}
Boyd, S. and Vandenberghe, L.
\newblock \emph{Convex optimization}.
\newblock Cambridge university press, 2004.

\bibitem[Brock et~al.(2019)Brock, Donahue, and Simonyan]{brock2018large}
Brock, A., Donahue, J., and Simonyan, K.
\newblock Large scale {GAN} training for high fidelity natural image synthesis.
\newblock In \emph{International Conference on Learning Representations}, 2019.

\bibitem[Cao et~al.(2018)Cao, Ding, Lui, and Huang]{BRE2018}
Cao, Y., Ding, G.~W., Lui, K. Y.-C., and Huang, R.
\newblock Improving {GAN} training via binarized representation entropy (bre)
  regularization.
\newblock \emph{ICLR}, 2018.
\newblock accepted as poster.

\bibitem[Cover \& Thomas(2012)Cover and Thomas]{cover2012elements}
Cover, T.~M. and Thomas, J.~A.
\newblock \emph{Elements of information theory}.
\newblock John Wiley \& Sons, 2012.

\bibitem[Danihelka et~al.(2017)Danihelka, Lakshminarayanan, Uria, Wierstra, and
  Dayan]{danihelka2017comparison}
Danihelka, I., Lakshminarayanan, B., Uria, B., Wierstra, D., and Dayan, P.
\newblock Comparison of maximum likelihood and gan-based training of real nvps.
\newblock \emph{arXiv preprint arXiv:1705.05263}, 2017.

\bibitem[Domke \& Sheldon(2018)Domke and Sheldon]{domke2018importance}
Domke, J. and Sheldon, D.~R.
\newblock Importance weighting and variational inference.
\newblock In \emph{Advances in neural information processing systems}, pp.\
  4470--4479, 2018.

\bibitem[Frey \& Hinton(1996)Frey and Hinton]{frey1996free}
Frey, B.~J. and Hinton, G.~E.
\newblock Free energy coding.
\newblock In \emph{Proceedings of Data Compression Conference-DCC'96}, pp.\
  73--81. IEEE, 1996.

\bibitem[Goodfellow et~al.(2014)Goodfellow, Pouget-Abadie, Mirza, Xu,
  Warde-Farley, Ozair, Courville, and Bengio]{goodfellow2014generative}
Goodfellow, I., Pouget-Abadie, J., Mirza, M., Xu, B., Warde-Farley, D., Ozair,
  S., Courville, A., and Bengio, Y.
\newblock Generative adversarial nets.
\newblock In \emph{Advances in neural information processing systems}, pp.\
  2672--2680, 2014.

\bibitem[Grosse et~al.(2015)Grosse, Ghahramani, and
  Adams]{grosse2015sandwiching}
Grosse, R.~B., Ghahramani, Z., and Adams, R.~P.
\newblock Sandwiching the marginal likelihood using bidirectional monte carlo.
\newblock \emph{arXiv preprint arXiv:1511.02543}, 2015.

\bibitem[Grover et~al.(2018)Grover, Dhar, and Ermon]{grover2018flow}
Grover, A., Dhar, M., and Ermon, S.
\newblock Flow-{GAN}: Combining maximum likelihood and adversarial learning in
  generative models.
\newblock In \emph{Thirty-Second AAAI Conference on Artificial Intelligence},
  2018.

\bibitem[Gulrajani et~al.(2017)Gulrajani, Ahmed, Arjovsky, Dumoulin, and
  Courville]{gulrajani2017improved}
Gulrajani, I., Ahmed, F., Arjovsky, M., Dumoulin, V., and Courville, A.~C.
\newblock Improved training of wasserstein {GAN}s.
\newblock In \emph{Advances in Neural Information Processing Systems}, pp.\
  5767--5777, 2017.

\bibitem[Heusel et~al.(2017)Heusel, Ramsauer, Unterthiner, Nessler, and
  Hochreiter]{fid}
Heusel, M., Ramsauer, H., Unterthiner, T., Nessler, B., and Hochreiter, S.
\newblock {GAN}s trained by a two time-scale update rule converge to a local
  nash equilibrium.
\newblock In Guyon, I., Luxburg, U.~V., Bengio, S., Wallach, H., Fergus, R.,
  Vishwanathan, S., and Garnett, R. (eds.), \emph{Advances in Neural
  Information Processing Systems 30}, pp.\  6626--6637. Curran Associates,
  Inc., 2017.

\bibitem[Hinton \& Van~Camp(1993)Hinton and Van~Camp]{hinton1993keeping}
Hinton, G. and Van~Camp, D.
\newblock Keeping neural networks simple by minimizing the description length
  of the weights.
\newblock In \emph{in Proc. of the 6th Ann. ACM Conf. on Computational Learning
  Theory}. Citeseer, 1993.

\bibitem[Huang et~al.(2018)Huang, Yuan, Xu, Guo, Sun, Wu, and
  Weinberger]{huang2018an}
Huang, G., Yuan, Y., Xu, Q., Guo, C., Sun, Y., Wu, F., and Weinberger, K.
\newblock An empirical study on evaluation metrics of generative adversarial
  networks, 2018.

\bibitem[Johnson et~al.(2016)Johnson, Alahi, and
  Fei-Fei]{johnson2016perceptual}
Johnson, J., Alahi, A., and Fei-Fei, L.
\newblock Perceptual losses for real-time style transfer and super-resolution.
\newblock In \emph{European conference on computer vision}, pp.\  694--711.
  Springer, 2016.

\bibitem[Karras et~al.(2018{\natexlab{a}})Karras, Aila, Laine, and
  Lehtinen]{karras2018progressive}
Karras, T., Aila, T., Laine, S., and Lehtinen, J.
\newblock Progressive growing of {GAN}s for improved quality, stability, and
  variation.
\newblock In \emph{International Conference on Learning Representations},
  2018{\natexlab{a}}.

\bibitem[Karras et~al.(2018{\natexlab{b}})Karras, Laine, and
  Aila]{karras2018style}
Karras, T., Laine, S., and Aila, T.
\newblock A style-based generator architecture for generative adversarial
  networks.
\newblock \emph{arXiv preprint arXiv:1812.04948}, 2018{\natexlab{b}}.

\bibitem[Kingma \& Ba(2014)Kingma and Ba]{adam}
Kingma, D.~P. and Ba, J.
\newblock Adam: {A} method for stochastic optimization.
\newblock \emph{CoRR}, abs/1412.6980, 2014.

\bibitem[Kingma \& Welling(2013)Kingma and Welling]{kingma2013auto}
Kingma, D.~P. and Welling, M.
\newblock Auto-encoding variational bayes.
\newblock \emph{arXiv preprint arXiv:1312.6114}, 2013.

\bibitem[Kingma et~al.(2019)Kingma, Abbeel, and Ho]{kingma2019bit}
Kingma, F.~H., Abbeel, P., and Ho, J.
\newblock Bit-swap: Recursive bits-back coding for lossless compression with
  hierarchical latent variables.
\newblock \emph{arXiv preprint arXiv:1905.06845}, 2019.

\bibitem[Krizhevsky \& Hinton(2009)Krizhevsky and
  Hinton]{krizhevsky2009learning}
Krizhevsky, A. and Hinton, G.
\newblock Learning multiple layers of features from tiny images.
\newblock Technical report, Citeseer, 2009.

\bibitem[LeCun et~al.(1998)LeCun, Bottou, Bengio, Haffner,
  et~al.]{lecun1998gradient}
LeCun, Y., Bottou, L., Bengio, Y., Haffner, P., et~al.
\newblock Gradient-based learning applied to document recognition.
\newblock \emph{Proceedings of the IEEE}, 86\penalty0 (11):\penalty0
  2278--2324, 1998.

\bibitem[Makhzani et~al.(2015)Makhzani, Shlens, Jaitly, Goodfellow, and
  Frey]{makhzani2015adversarial}
Makhzani, A., Shlens, J., Jaitly, N., Goodfellow, I., and Frey, B.
\newblock Adversarial autoencoders.
\newblock \emph{arXiv preprint arXiv:1511.05644}, 2015.

\bibitem[Miyato et~al.(2018)Miyato, Kataoka, Koyama, and
  Yoshida]{miyato2018spectral}
Miyato, T., Kataoka, T., Koyama, M., and Yoshida, Y.
\newblock Spectral normalization for generative adversarial networks.
\newblock \emph{arXiv preprint arXiv:1802.05957}, 2018.

\bibitem[Neal(2001)]{neal2001annealed}
Neal, R.~M.
\newblock Annealed importance sampling.
\newblock \emph{Statistics and computing}, 11\penalty0 (2):\penalty0 125--139,
  2001.

\bibitem[Neal(2005)]{neal2005estimating}
Neal, R.~M.
\newblock Estimating ratios of normalizing constants using linked importance
  sampling.
\newblock \emph{arXiv preprint math/0511216}, 2005.

\bibitem[Neal et~al.(2011)]{neal2011mcmc}
Neal, R.~M. et~al.
\newblock Mcmc using hamiltonian dynamics.
\newblock \emph{Handbook of markov chain Monte Carlo}, 2\penalty0
  (11):\penalty0 2, 2011.

\bibitem[Radford et~al.(2015)Radford, Metz, and
  Chintala]{radford2015unsupervised}
Radford, A., Metz, L., and Chintala, S.
\newblock Unsupervised representation learning with deep convolutional
  generative adversarial networks.
\newblock \emph{arXiv preprint arXiv:1511.06434}, 2015.

\bibitem[Rezende \& Viola(2018)Rezende and Viola]{rezende2018taming}
Rezende, D.~J. and Viola, F.
\newblock Taming vaes.
\newblock \emph{arXiv preprint arXiv:1810.00597}, 2018.

\bibitem[Sajjadi et~al.(2018)Sajjadi, Bachem, Lucic, Bousquet, and
  Gelly]{sajjadi2018assessing}
Sajjadi, M.~S., Bachem, O., Lucic, M., Bousquet, O., and Gelly, S.
\newblock Assessing generative models via precision and recall.
\newblock In \emph{Advances in Neural Information Processing Systems}, pp.\
  5228--5237, 2018.

\bibitem[Salimans et~al.(2016)Salimans, Goodfellow, Zaremba, Cheung, Radford,
  and Chen]{salimans2016improved}
Salimans, T., Goodfellow, I., Zaremba, W., Cheung, V., Radford, A., and Chen,
  X.
\newblock Improved techniques for training gans.
\newblock In \emph{Advances in Neural Information Processing Systems}, pp.\
  2234--2242, 2016.

\bibitem[Salimans et~al.(2018)Salimans, Zhang, Radford, and
  Metaxas]{salimans2018improving}
Salimans, T., Zhang, H., Radford, A., and Metaxas, D.
\newblock Improving {GAN}s using optimal transport.
\newblock In \emph{International Conference on Learning Representations}, 2018.

\bibitem[Theis et~al.(2015)Theis, Oord, and Bethge]{theis2015note}
Theis, L., Oord, A. v.~d., and Bethge, M.
\newblock A note on the evaluation of generative models.
\newblock \emph{arXiv preprint arXiv:1511.01844}, 2015.

\bibitem[Theis et~al.(2017)Theis, Shi, Cunningham, and
  Husz{\'a}r]{theis2017lossy}
Theis, L., Shi, W., Cunningham, A., and Husz{\'a}r, F.
\newblock Lossy image compression with compressive autoencoders.
\newblock \emph{arXiv preprint arXiv:1703.00395}, 2017.

\bibitem[Townsend et~al.(2019)Townsend, Bird, and
  Barber]{townsend2019practical}
Townsend, J., Bird, T., and Barber, D.
\newblock Practical lossless compression with latent variables using bits back
  coding.
\newblock \emph{arXiv preprint arXiv:1901.04866}, 2019.

\bibitem[Wallace(1990)]{wallace1990classification}
Wallace, C.~S.
\newblock Classification by minimum-message-length inference.
\newblock In \emph{International Conference on Computing and Information}, pp.\
   72--81. Springer, 1990.

\bibitem[Wang et~al.(2003)Wang, Simoncelli, and Bovik]{wang2003multiscale}
Wang, Z., Simoncelli, E.~P., and Bovik, A.~C.
\newblock Multiscale structural similarity for image quality assessment.
\newblock In \emph{The Thrity-Seventh Asilomar Conference on Signals, Systems
  \& Computers, 2003}, volume~2, pp.\  1398--1402. Ieee, 2003.

\bibitem[Wang et~al.(2004)Wang, Bovik, Sheikh, Simoncelli,
  et~al.]{wang2004image}
Wang, Z., Bovik, A.~C., Sheikh, H.~R., Simoncelli, E.~P., et~al.
\newblock Image quality assessment: from error visibility to structural
  similarity.
\newblock \emph{IEEE transactions on image processing}, 13\penalty0
  (4):\penalty0 600--612, 2004.

\bibitem[Wu et~al.(2016)Wu, Burda, Salakhutdinov, and
  Grosse]{wu2016quantitative}
Wu, Y., Burda, Y., Salakhutdinov, R., and Grosse, R.
\newblock On the quantitative analysis of decoder-based generative models.
\newblock \emph{arXiv preprint arXiv:1611.04273}, 2016.

\bibitem[Yang et~al.(2020)Yang, Bamler, and Mandt]{yang2020improving}
Yang, Y., Bamler, R., and Mandt, S.
\newblock Improving inference for neural image compression.
\newblock \emph{arXiv preprint arXiv:2006.04240}, 2020.

\end{thebibliography}
